\RequirePackage{fix-cm}
\documentclass{arxiv/bytedance}
\usepackage{tabularx}
\usepackage{array}
\usepackage{enumitem}
\usepackage{float}
\usepackage{algorithm}
\usepackage{algpseudocode}
\usepackage{xspace}

\definecolor{gublue}{rgb}{0,0.66,1}

\definecolor{darkgreen}{RGB}{70,145,110}

\definecolor{siyupurple}{rgb}{0.55,0.15,0.75}

\newcommand{\tool}{RSIHub\xspace}
\title{RSIHub: A Composable Framework \\ for Agent-Evolution Methods}

\author[1, 4, \star]{Zimu Wang}
\author[2, 6, \star]{Xiaobo Wang}
\author[3, 7, \star]{Siyu Ye}
\author[6, \star]{Jiaqi Li}
\author[4]{Silin Chen}
\author[4]{Yuling Shi}
\author[3, 7]{\\ Ruobing Wang}
\author[1]{Jingjing Zhang}
\author[4]{Chaofan Wang}
\author[5]{Kai Wu}
\author[3, 7]{\\ Mengnan Qi}
\author[3, 7]{Kai Cai}
\author[4,\dagger]{Xiaodong Gu}
\author[6,\dagger]{Zilong Zheng}

\affiliation[1]{University of California, Berkeley}
\affiliation[2]{University of Science and Technology of China}
\affiliationnewline[3]{Douyin Group}
\affiliation[4]{Shanghai Jiao Tong University}
\affiliation[5]{Tongji University}
\affiliationnewline[6]{Beijing Institute for General Artificial Intelligence}
\affiliation[7]{Simple Agent Lab}
\contribution[\star]{Equal contribution}
\contribution[\dagger]{Corresponding author}

\abstract{
Agent-evolution methods use execution feedback to improve prompts, skills, tools, and orchestration, but their implementations often rebuild the same experimental machinery. We present \tool, an open-source framework for building, comparing, and extending these methods on shared infrastructure. Recipes compose reusable operators and declare mutable resources; \tool supplies execution, evaluation, recovery, and lineage. Researchers can focus on evolution policy while evaluation stays outside the mutable surface and results remain tied to candidates and evaluation conditions. We instantiate AHE, Hyperagents, A-Evolve, and GEPA without modifying the framework core, keeping method-specific logic in recipes and operators. We evaluate them with MiniSWE and Codex on Terminal Bench 2 and $\tau^3$ Banking under fixed model and evaluator configurations. The largest observed overall pass@1 gain is 13.5 percentage points, while method rankings vary across settings and optimization gains do not always transfer to held-out tasks. \tool provides a common protocol for comparing evolution policies, inspectable candidate histories, and reusable components for developing new methods.
}

\date{October 22, 2026}
\correspondence{Siyu Ye at \email{yesiyu.andy@bytedance.com}, Xiaodong Gu at \email{xiaodong.gu@sjtu.edu.cn}, Jiaqi Li at \email{lijiaqi@bigai.ai}, Zilong Zheng at \email{zlzheng@bigai.ai}, Zimu Wang at \email{zm.wang@berkeley.edu}}

\checkdata[Code]{\url{https://github.com/simple-agent-lab/RSIHub}}
\checkdata[Project Page]{\url{https://simpleagentlab.com/rsihub/}}

\hypersetup{pdftitle={RSIHub: A Composable Framework for Agent-Evolution Methods},
  pdfauthor={Zimu Wang and collaborators}}

\begin{document}
\raggedbottom
\maketitle

\section{Introduction}
\label{sec:introduction}

% \gu{what is the purpose of this paragraph?}\wang{done} 
Language-model agents are commonly developed through an
empirical loop: developers inspect executions, revise prompts, reusable skills,
tools, or orchestration, and reevaluate the resulting behavior
\cite{lin2026ahe,openhandsplatform}. Agent-evolution systems
automate parts of this loop by turning execution feedback into persistent
changes to the agent
\cite{lin2026ahe,agrawal2025gepa,zhang2026ace,zhang2026evoskills}; some also
revise parts of the improvement procedure
\cite{fernando2024promptbreeder,zhang2026hyperagents,zhang2025dgm}.
Program-optimization methods apply a similar feedback loop to language-model
programs and textual components
\cite{khattab2023dspy,yuksekgonul2024textgrad}. Across both settings, each
iteration selects a candidate to extend, gathers evidence from its behavior,
proposes and evaluates a revision, and retains the outcome for later search.
This recurring lifecycle provides a common structure across these systems.

Agent-evolution methods vary in how they instantiate this lifecycle and in which resources they allow to change.
Much of the supporting infrastructure nevertheless recurs: isolated candidate
workspaces, task execution and evaluation, artifact retention, recovery, and lineage.
Implementations commonly bundle this infrastructure with method policy. As a
result, implementing a method can require building another end-to-end system, while
comparing methods requires reconciling runners, configurations, and evaluator
interfaces. These costs motivate common infrastructure on which researchers can test new
evolution ideas without rebuilding workspaces, execution, and evaluation.
% \xiaobo{This sentence can be more high-level. "A unified framework is essential for promoting agent evolution. xxx"}\wang{done}

\begin{figure}[!t]
  \centering
  \includegraphics[width=0.9\linewidth]{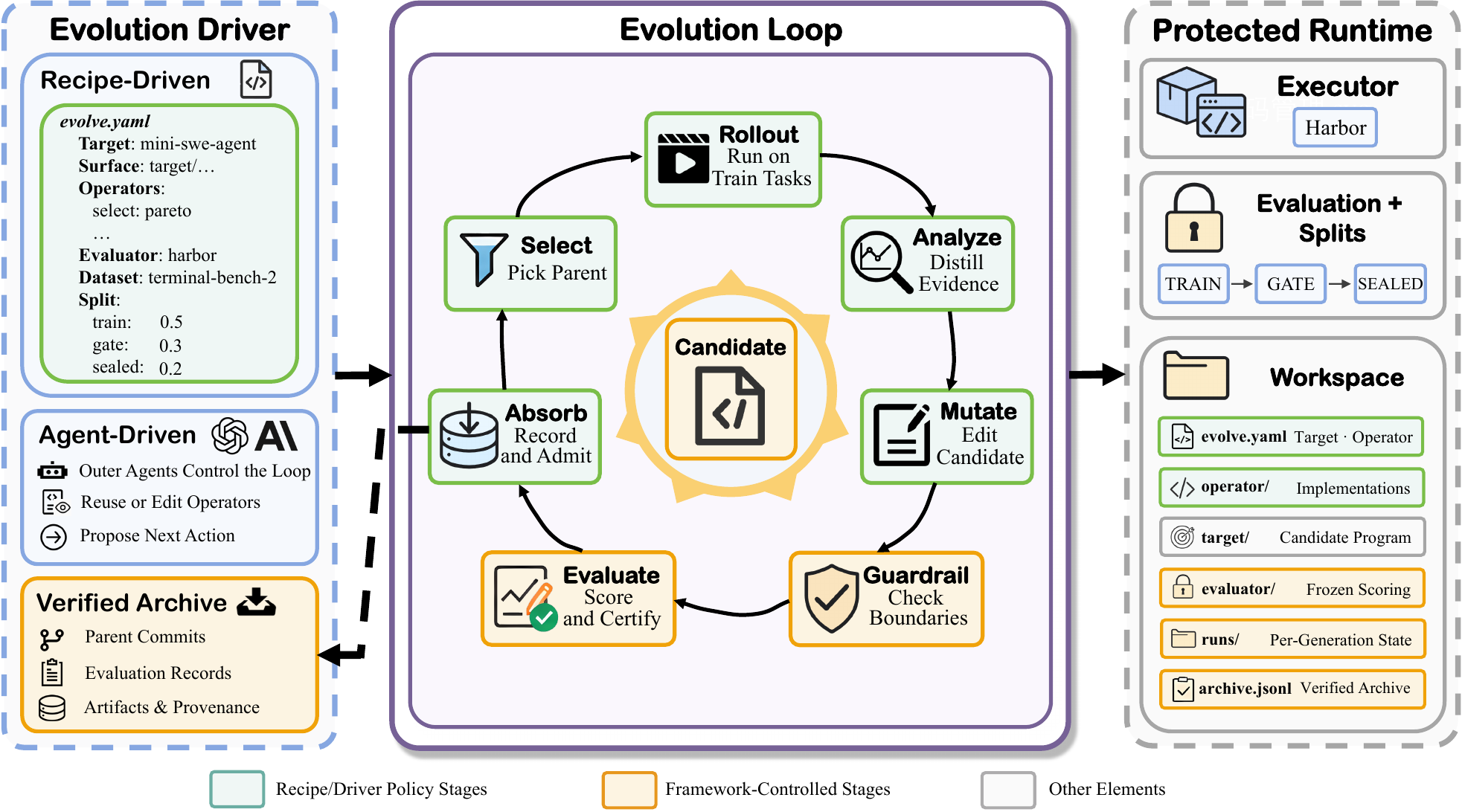}
  \caption{\tool separates method policy from framework-managed execution. In a
  recipe-driven run, a researcher-authored recipe remains fixed across
  iterations. In an agent-driven run, an outer agent uses permitted feedback
  from each completed iteration to revise the recipe's operator bindings and
  policy settings for the next. At each iteration boundary, \tool validates and
  snapshots the active recipe before running the same guarded lifecycle. The
  target definition, evaluator, runtime, data splits, and mutable surface remain
  fixed; surface enforcement, protected evaluation, and verified archive writes
  stay under framework control.}
  \label{fig:integrity-architecture}
\end{figure}

To meet this need, we introduce \tool, a composable framework that separates
method policy from this recurring infrastructure. As
\Cref{fig:integrity-architecture} illustrates, a \emph{recipe} binds one
implementation to each enabled operator interface, declares the mutable
surface, and specifies the run configuration: the target, evaluator, runtime,
and data splits. A framework-managed \emph{evolution loop} invokes the selected operators
to choose a parent, gather evidence, revise a candidate, and decide what to
retain. A \emph{protected runtime} executes and evaluates candidates under that
configuration, and the framework writes their outcomes to the verified archive.
In the \emph{agent-driven} mode, an outer agent adapts the recipe between
iterations under the same lifecycle and evaluation boundaries.
Shared framework services provide workspace construction, artifacts, recovery,
and lineage. New operator implementations plug into these stable interfaces,
leaving the framework core unchanged. This follows the
organization used by research frameworks in
adjacent fields: stable interfaces expose research choices while shared runtime
mechanics support their execution \cite{paszke2019pytorch,sheng2025hybridflow}. 
% \xiaobo{Some definitions mismatch with the figure.}\wang{done, and the figure will be updated soon by silin}

The protected-runtime layer addresses a central attribution risk in broadly
mutable evolution systems. Method-controlled
changes to evaluation can alter measurement conditions, while changes to record
keeping can rewrite the reported evidence; either can make an apparent gain
ambiguous.
\tool addresses this risk by keeping evaluation and its archive under
framework control, outside the declared mutable surface. Framework services
associate each result with the committed candidate and evaluation conditions
that produced it; a separate Gate operator decides whether the evaluated
candidate is eligible to become a parent.

Our evaluation compares recipes for four methods across two seed agents and two
benchmarks under fixed model and evaluator configurations. It examines
effectiveness and held-out transfer, execution fidelity, integrity through
interruption recovery and boundary enforcement, and runtime and implementation
cost.

Concretely, this paper makes three contributions:
\begin{itemize}[leftmargin=*,itemsep=2pt,topsep=3pt]
  \item \textbf{Composable recipes.} We introduce a recipe abstraction that
        specifies method policy through reusable lifecycle operators and
        declares the resources that may be revised. We express AHE,
        Hyperagents, A-Evolve, and GEPA as recipes over the same operators.
        % \xiaobo{This item is somehow similar to the first item and can be merged into it.}\siyu{done: the former third item (recipes for the four methods) is merged here}
  \item \textbf{Shared experimental infrastructure.} We provide a common
        lifecycle for candidate execution, recovery, artifacts, certified
        evaluation records, and lineage.
  \item \textbf{Evaluation on common ground.} We compare the four recipes
        across two seed agents and two benchmarks under fixed model and
        evaluator configurations. We also test the framework itself: whether
        it changes the measured capability of an unevolved agent, whether it
        recovers from interruption and holds its boundaries, and what it costs
        in runtime and code. Rankings among the recipes vary across agents,
        benchmarks, and splits.
        % \xiaobo{\textbf{Comprehensive framework evaluation.} We conduct a comprehensive evaluation of \tool across two benchmarks, two seed agents, and four evolution methods, examining not only evolution effectiveness and transfer across task splits, but also capability preservation, recovery from interruption, integrity enforcement, runtime overhead, and integration effort.}\siyu{done: folded into this item; kept the three framework tests, dropped "comprehensive" and "not only / but also"}
\end{itemize}

\paragraph{Open-source resources.} We release \tool through Simple Agent Lab under Apache-2.0, with runnable recipes, reusable operators, documentation\footnote{\url{https://simpleagentlab.com/RSIHub/}}, and an experiment viewer. Researchers can start from runnable recipes, inspect candidate changes and evaluation results, and contribute new recipes and operators.

\section{Related Work}
\label{sec:related}

\paragraph{Agent optimization and evolution.} Agent improvement spans inference-time feedback, persistent artifact optimization, and self-modification. Reflexion retains linguistic feedback across trials; Self-Refine revises outputs at test time~\cite{shinn2023reflexion,madaan2023selfrefine}. RAM improves external memory through communications~\cite{li2024ram}. Related work examines memory mechanisms~\cite{jia2025aihippocampus}, reflection learning~\cite{li2025reflectevo}, uncertainty-aware preference optimization~\cite{wang2025uapo}, and reward-model self-improvement~\cite{wang2026flip}. DSPy and TextGrad optimize language-model programs or textual variables~\cite{khattab2023dspy,yuksekgonul2024textgrad}. Promptbreeder evolves task and mutation prompts; ACE evolves context playbooks; and CoEvoSkills evolves multi-file skills with a separate surrogate verifier~\cite{fernando2024promptbreeder,zhang2026ace,zhang2026evoskills}. EffiSkill mines reusable skills for code-efficiency optimization~\cite{wang2026effiskill}. GPTSwarm optimizes agent graphs, while AgentSquare searches combinations of planning, reasoning, tool-use, and memory modules~\cite{zhuge2024gptswarm,shang2025agentsquare}. EvoAgentX integrates optimizers for evolving multi-agent workflows~\cite{wang2025evoagentx}. In autoresearch, Fork-and-Flush combines workspace inheritance with fresh conversational contexts and branch selection to diversify candidate search~\cite{cai2026forkandflush}. Our recipes adapt selected policies from four approaches. A-Evolve frames deployment-time adaptation as goal-directed optimization over persistent system state~\cite{lin2026aevolve}. AHE guides revisions to harness files with distilled execution evidence~\cite{lin2026ahe}; GEPA uses trajectory reflection and Pareto selection to optimize prompts~\cite{agrawal2025gepa}; and Hyperagents combines a task agent and a self-modifying meta agent in one editable program, allowing the improvement procedure to evolve~\cite{zhang2026hyperagents}. Other approaches include STOP, which revises optimizer scaffolding; ADAS, which generates agent programs with a meta agent; DGM, which modifies its coding agent; and AlphaEvolve, which evolves target algorithms~\cite{zelikman2024stop,hu2024adas,zhang2025dgm,novikov2025alphaevolve}. \tool provides shared execution, lifecycle, and evidence services for comparing our four policy adaptations.

\paragraph{Tasks, environments, and execution infrastructure.}
Benchmarks cover software engineering,
terminal interaction, and knowledge-grounded
dialogue~\cite{jimenez2024swebench,merrill2026terminalbench,shi2026tauknowledge}.
Others examine multilingual code refactoring~\cite{shi2026swebenchpromax},
repository exploration~\cite{zhang2026sweexplore}, and repository-level question
answering~\cite{peng2025sweqa}.
Harbor standardizes task, dataset, agent, and container-environment interfaces
for evaluation and optimization~\cite{harbor2026framework}. The OpenHands
Software Agent SDK provides composable agent, tool, conversation, and workspace
abstractions with local or remote execution~\cite{openhandsplatform}. \tool
delegates container execution, trajectory capture, and per-task verification to
Harbor, and adds a versioned evolution lifecycle, recipe-level method
composition, and common provenance across methods.

\paragraph{Provenance and measurement validity.}
The AI4AI survey distinguishes measured gains, retention, and held-out transfer~\cite{wu2026ai4ai}.
Software-provenance and reproducibility systems authenticate authorized
production steps, bind identities to signed artifacts, or package executions
with their
dependencies~\cite{torresarias2019intoto,newman2022sigstore,chirigati2016reprozip}.
\tool records candidate lineage,
distinguishes parent selection from post-evaluation Gate decisions, and binds
outcomes to candidates, recipes, and evaluation conditions.
Controlled skill-evolution experiments show that persistent improvement can be
sparse and depend on feedback, validation, models, and benchmarks~\cite{liu2026rethinking}. In a synthetic reward-model setting, overoptimizing an
imperfect proxy can reduce performance under a fixed gold-standard
reward model~\cite{gao2022overoptimization}. After curriculum training in
deliberately gameable environments, behavior meeting an operational definition
of reward tampering appeared in a small but non-negligible share of cases; the
authors note that this definition admits false
positives~\cite{denison2024subterfuge}. \tool services write certified evaluation fields
and filter operator-supplied Record annotations before appending them. This supports attribution under the framework's trust
assumptions; benchmark overfitting and broader specification gaming remain
evaluator- and study-level concerns.

% Shared with the main paper.
\section{The RSIHub Framework}
\label{sec:rsihub}
%\gu{can we merge section 3 and 4?}\wang{done}

\tool separates method policy from the shared mechanics of candidate execution,
evaluation, and record keeping. This section states the principles behind that
separation, then describes how the candidate lifecycle and provenance model
realize it.

\subsection{Design principles}
\label{sec:principles}

RSIHub is built so that a researcher can assemble an evolution method from
interchangeable parts, decide what those parts may change and what they may
see, and reuse framework services for recurring mechanics.

\textbf{Assemble methods from interchangeable parts.}
RSIHub fixes the slots of the evolution lifecycle and leaves their contents
open. A \emph{stage} is a slot such as \texttt{select} or \texttt{mutate}. An
\emph{operator} is one implementation of a stage. A \emph{recipe} names the
operator it wants in each slot and configures it. When the required operator
implementations are available, a recipe can express the method-specific choices
through configuration.

\textbf{Say what a method may change and see.}
Two recipe declarations make this an experimental choice. The
\texttt{surface} block declares the resources a candidate may change, allowing
the same operator interfaces to support prompt-only and whole-harness studies.
The \texttt{mutate} block configures how the editing agent runs and which
retained context it receives. Together, these declarations distinguish the persistent resources being evolved from the information used to propose each revision.

\textbf{Provide the rest of the pipeline once.}
Workspace construction, candidate snapshots, evaluation execution, artifact
retention, terminal-state classification, recovery from interrupted runs, and
lineage bookkeeping commonly recur across evolution methods. RSIHub implements
these services once and gives every recipe the same execution path. When a
study changes one operator without changing the other recipe fields, it isolates
one declared policy component while reusing the same framework mechanics.

\textbf{Keep scoring out of the method's reach.}
Method policy may revise candidates and the operators that propose them.
Measurement remains a framework responsibility: framework services
own evaluation and the evidence record outside the mutable surface a recipe declares,
while parent eligibility remains a separate downstream policy choice.

Together, these principles separate method-specific choices from
framework-managed execution and measurement. The candidate lifecycle below
makes this separation operational.

\subsection{Composable candidate lifecycle}

The lifecycle follows the seven phases shown in
\Cref{fig:integrity-architecture}: Select, Rollout, Analyze, Mutate, Guardrail,
Evaluate, and Absorb. Across these phases, recipes bind five required operator
interfaces---Select, Rollout, Mutate, Gate, and Record---and may bind four
optional ones, marked with $^*$ below: Analyze, Validate, Novelty, and Reflect.

\tool implements this lifecycle on Harbor~\cite{harbor2026framework}, which
provides container execution, trajectory capture, and per-task verification.
Above this substrate, \tool manages candidate lineage, train/gate/sealed splits,
mutable-surface enforcement, and certified evaluation records.

The lifecycle accepts either a fixed, researcher-authored recipe or one supplied
by an outer agent. In the agent-driven mode, the outer agent uses a
recipe-authoring skill to compose the initial recipe and may revise its operator
bindings and policy settings using permitted feedback after each completed
iteration. The outer agent adapts the evolution procedure, while Mutate edits
the candidate within an iteration. The target, evaluator, runtime, data splits,
and allowed mutable surface remain fixed; sealed tasks are excluded from
adaptation feedback. Our experiments evaluate fixed recipes.

Before each iteration, the framework validates and snapshots the active recipe,
resolving every binding and checking its settings against the operator's schema.
Each operator runs in a separate process, and the framework validates its
structured output before using it.

The archive separates framework-certified fields, the evaluated candidate, its
parent, the evaluation conditions, and the outcome, from method-specific
annotations. Operators propose annotations; the framework filters and writes
them.

\begin{itemize}[leftmargin=*,itemsep=1pt,topsep=3pt,parsep=0pt]
  \item \textbf{Select.} Chooses an eligible parent from the archive according
        to the recipe's search policy. The framework verifies its identity and
        forks its exact recorded commit.
  \item \textbf{Rollout.} Runs the parent on training tasks and returns
        execution evidence as a summary and retained artifact references. The
        supplied Harbor rollout uses only the training split; a replay-based
        rollout reuses the parent's recorded evaluation instead of running
        again.
  \item \textbf{Analyze\textsuperscript{*}.} Distills rollout cases into
        bounded mutation evidence, such as failure patterns or compact traces,
        in the representation Mutate consumes.
  \item \textbf{Mutate.} Proposes edits within the declared mutable surface.
        The operator is typically an agent rather than a fixed transformation:
        it works over multiple turns, may run small-scale trials on training
        tasks before proposing a change, and, as in Hyperagents, may edit the
        evolution operators themselves when the surface allows. It reports
        changed paths, notes, and usage.
  \item \textbf{Guardrail.} The framework checks the actual tree diff against
        the mutable surface. Two optional operators may then reject the
        proposal: Validate\textsuperscript{*} applies a method-specific
        pre-evaluation check, and Novelty\textsuperscript{*} drops near-duplicates
        of accepted diffs before they consume evaluation budget.
  \item \textbf{Evaluate.} This framework-managed phase commits the reviewed
        tree, scores a clean checkout on the gate tasks, and certifies the result.
        Gate tasks may coincide with training tasks; sealed tasks are reserved
        for held-out reporting.
  \item \textbf{Absorb.} Gate decides from the certified score whether the
        child becomes a parent, Record proposes method-specific annotations, and
        Reflect\textsuperscript{*} distills insights from archive history for
        later generations. The framework records permitted outputs; none of
        these operators can change certified evaluation fields.
\end{itemize}

\Cref{alg:candidate-transaction} expands these phases into their operator calls
and framework actions over the validated recipe snapshot $R$ for that iteration
and archive history $H$.
The proposal $c$ remains uncommitted
through Guardrail; $\hat{c}$ is the clean commit used for protected evaluation.
In the pseudocode, $\Call{Stops}{x}$ denotes rejection, failure, malformed
output, or timeout. An evaluation is \emph{benchmark-complete}
when its trial outcomes satisfy the configured coverage policy; only then does
Gate run.

A skipped Analyze passes through $X$, and skipped Validate
and Novelty accept. Configured Reflect appends validated operations to a separate
playbook without modifying $H$. Empty proposals and surface
violations likewise close as typed, archived attempts.
Failures of Rollout, Analyze, or Mutate take the same
$\Call{CloseAndRecord}{}$ path and are elided for compactness.

\begin{algorithm}
  \caption{One RSIHub candidate transaction.}
  \label{alg:candidate-transaction}
  \footnotesize
  \newcommand{\lightalgkeyword}[1]{{\fontseries{b}\selectfont\color{black!92}#1}}
  \algrenewcommand\algorithmicend{\lightalgkeyword{end}}
  \algrenewcommand\algorithmicprocedure{\lightalgkeyword{procedure}}
  \algrenewcommand\algorithmicif{\lightalgkeyword{if}}
  \algrenewcommand\algorithmicthen{\lightalgkeyword{then}}
  \algrenewcommand\algorithmicelse{\lightalgkeyword{else}}
  \algrenewcommand\algorithmicreturn{\lightalgkeyword{return}}
  \begin{minipage}[t]{0.49\linewidth}
    \vspace{0pt}
    \begin{algorithmic}[1]
      \Procedure{EvolveOne}{$R,H$}
        \State $p \gets \Call{Select}{H}$
        \State $c \gets \Call{ForkCertifiedParent}{p,H}$
        \State $X \gets \Call{Rollout}{c}$
        \State $Z \gets \Call{Analyze}{c,X}^{*}$
        \State \Call{Mutate}{$c,Z$}
        \State $s \gets \Call{CheckSurface}{p,c,R_{\mathrm{surface}}}$
        \If{$\Call{Stops}{s}$}
          \State \Return $\Call{CloseAndRecord}{H,p,c,s}$
        \EndIf
        \State $v \gets \Call{Validate}{c}^{*}$
        \If{$\Call{Stops}{v}$}
          \State \Return $\Call{CloseAndRecord}{H,p,c,v}$
        \EndIf
        \State $n \gets \Call{Novelty}{p,c,H}^{*}$
        \If{$\Call{Stops}{n}$}
          \State \Return $\Call{CloseAndRecord}{H,p,c,n}$
        \EndIf
        \algstore{candidate}
    \end{algorithmic}
  \end{minipage}\hfill
  \begin{minipage}[t]{0.49\linewidth}
    \vspace{0pt}
    \begin{algorithmic}[1]
        \algrestore{candidate}
        \State $\hat{c} \gets \Call{CommitReviewedTree}{c}$
        \State $q \gets \Call{ResolveContract}{R,\hat{c}}$
        \State $y \gets \Call{Evaluate}{\hat{c},q}$
        \State $H \gets \Call{Certify}{H,p,\hat{c},q,y}$
        \If{$y.\mathrm{outcome} \ne \mathrm{benchmark\_complete}$}
          \State \Return $H$
        \EndIf
        \State $a \gets \Call{Gate}{p,\hat{c},y,H}$
        \If{$a.\mathrm{failed}$}
          \State $H \gets \Call{CloseAndRecord}{H,p,\hat{c},a}$
        \Else
          \State $H \gets \Call{AppendGateDecision}{H,a}$
          \State $e \gets \Call{Record}{p,\hat{c},y,a,H}$
          \State $H \gets \Call{AppendAllowedRecord}{H,e}$
        \EndIf
        \State $\Call{Reflect}{H}^{*}$
        \State \Return $H$
      \EndProcedure
    \end{algorithmic}
  \end{minipage}
\end{algorithm}

When a recipe places operators inside the mutable surface, the framework must
decide which version of an operator runs. This candidate-level mechanism is
distinct from an outer agent revising recipe bindings or settings between
iterations. For each stage invocation, it loads the
implementation from the checkout supplied to that stage. Changes to already-run
stages take effect in a later generation,
while changes to Validate or Novelty may affect the current proposal. Gate and
Record subsequently load from the committed child. The certified record provides a recovery checkpoint after evaluation: it holds the committed child and its outcome before downstream finalization. Gate, Record, and Reflect may add policy outputs but cannot alter certified evaluation fields.

\subsection{Provenance}

A certified evaluation event ties the committed candidate, its parent, and the
active recipe snapshot to evaluator and task-set fingerprints, trials, score, costs,
artifact references, and lineage. A hash-referenced contract can also record evaluator-configuration and dataset-content digests,
split and repetition settings, model identity, retry policy, evaluator-runtime
digest, and framework version. This identity lets a study distinguish a change
in method policy from a change in recorded evaluation conditions.

\paragraph{From experiment records to method design.} Researchers can start from a bundled recipe and inspect candidate snapshots, lineage, task outcomes, costs, and trajectories to investigate stalled progress and revision patterns, combining artifacts with scores as in Fork-and-Flush~\cite{cai2026forkandflush}. For example, researchers can replace the parent-selection operator in a separate run while holding the seed agent, mutable surface, model, evaluator, and task splits fixed. \tool supplies execution, recovery, and provenance services, with command-line summaries and a read-only viewer for inspecting the resulting records (\Cref{app:execution-viewer}). This workflow lets researchers test new policy components while reusing the surrounding experimental infrastructure. At the evaluated commit, the four adaptations used 47--61 recipe lines and reused two operator bindings each, with no framework-source changes (\Cref{tab:integration-effort}). Method-specific operator code ranged from 216 to 1,087 lines.

% Code evidence: docs/concepts/design.md, src/evolve/frozen/interfaces.py,
% driver.py, surface.py, candidate/snapshot.py, evaluation/contract.py,
% evaluation/execution.py, evaluation/results.py, archive.py, feedback.py.

\section{Evaluation}
\label{sec:evaluation}

In this section, we organize the evaluation around three research questions:
\begin{itemize}
[leftmargin=*,itemsep=1pt,topsep=2pt]
  \item \textbf{RQ1:} Can \tool support effective agent evolution?
  \item \textbf{RQ2:} Does \tool preserve the capabilities of the agents it executes?
  \item \textbf{RQ3:} Does \tool preserve the integrity of agent-evolution experiments?
\end{itemize}

To answer these questions, we compare evolution recipes across agents, benchmarks, and held-out splits for RQ1, use generation-zero controls against direct execution for RQ2, and conduct fault-injection and boundary-enforcement trials for RQ3.

\subsection{Recipe instantiations} \label{sec:recipe}

\tool includes four recipes that adapt selected policies from A-Evolve~\cite{lin2026aevolve}, AHE~\cite{lin2026ahe}, GEPA~\cite{agrawal2025gepa}, and Hyperagents~\cite{zhang2026hyperagents}; a hill-climbing recipe provides a minimal baseline. Their operator configurations and mutation targets illustrate how the shared execution path supports different method policies.

\begin{center}
\small
\begin{tabular}{@{}lll@{}}
\toprule
Recipe & Search behavior & Mutation focus \\
\midrule
\texttt{hill\_climb}  & continue from the best parent      & target \\
\texttt{aevolve}      & goal-directed persistent adaptation & prompt, target skills \\
\texttt{ahe}          & harness revision from execution evidence & target \\
\texttt{gepa}         & Pareto selection; minibatch validation & prompt, task skill \\
\texttt{hyperagents}  & co-evolve target and evolution policy & target, selected operators \\
\bottomrule
\end{tabular}
\end{center}

Each recipe is an interface adaptation at the level of operator choices and configuration. Exact correspondence with a source implementation also depends on its prompts, models, budgets, dependencies, and runtime adapters. A user can run an adaptation unchanged, swap one operator, revise its mutable resources, or write a new recipe over the same services.

\subsection{Setup}

\textbf{Agents and evolution methods.} We evolve MiniSWE \cite{yang2024sweagent} and Codex \cite{openai_codex} as seed agents with AHE~\cite{lin2026ahe}, Hyperagents~\cite{zhang2026hyperagents}, A-Evolve~\cite{lin2026aevolve}, and GEPA~\cite{agrawal2025gepa}. Each adaptation is instantiated as an \tool recipe over the same frozen Harbor evaluator~\cite{harbor2026framework}. The recipes retain their intended mutable surfaces as in \Cref{sec:recipe}.

\textbf{Benchmarks and metrics.} We evaluate the seed agent and evolved agent on Terminal Bench 2~\cite{merrill2026terminalbench} and $\tau^3$ Banking, which is introduced as the knowledge-grounded $\tau$-Banking domain by $\tau$-Knowledge~\cite{shi2026tauknowledge}, for one trial per task, and report pass@1. Terminal Bench 2 contains 89 tasks split into 50 train, 19 gate, and 20 sealed tasks; $\tau^3$ Banking contains 97 tasks split into 50 train, 20 gate, and 27 sealed tasks. Train trajectories provide evolution feedback, gate tasks support candidate admission, and sealed tasks are excluded from mutation feedback and used only for held-out reporting. Because the splits have different sizes, each split column in \Cref{tab:evolution-results} is its contribution to the full-benchmark score, e.g., $N_{\mathrm{pass},s}/N_{\mathrm{all}}$.

\textbf{Models.} All experiments use GPT-5.4 with high reasoning effort as the target model for both MiniSWE and Codex. Every recipe uses a Codex mutation agent backed by GPT-5.4 with xhigh reasoning effort. We hold the target and mutation model configurations fixed across benchmarks, seed agents, and evolution methods so that performance differences are attributable to the evolution recipes under shared model configurations.

\textbf{Optimization and runtime.} All recipes propose one child per generation, and evolution runs continue for at most 10 generations. The evolution results use static task sampling and one evaluation per task, while the capability-preservation controls use three independent repetitions. All task execution uses Docker-isolated Harbor environments. The setup script downloads and verifies content-pinned benchmark assets and builds the recipe-specific mutation container, pinning MiniSWE 2.4.5 or Codex 0.143.0.

% \Cref{tab:evolution-results} reports pass@1 before and after agent evolution. Each seed baseline appears on its own row; method rows show the final score and its change from that baseline.

\subsection{Evolution effectiveness and transfer}

\textbf{Protocol.} We test the effectiveness conjecture by comparing each evolved recipe with its seed under shared conditions. Overall score changes probe whether the framework can support effective improvement; train, gate, and sealed results reveal whether those improvements survive admission and extend beyond the trajectories used to propose revisions. Crossing agents, benchmarks, and methods makes the conjecture answerable across heterogeneous settings rather than a single favorable configuration.

\begin{table}[t]
  \centering
  \caption{Pass@1 before and after evolution on Terminal Bench 2 and $\tau^3$ Banking. All entries are percentages; split columns report their contribution to full-benchmark pass@1. Parenthesized values are absolute changes from the corresponding seed baseline, measured in percentage points. Among evolved variants, the best results per agent are in \textbf{bold} and the second best are \underline{underlined}.}
  \label{tab:evolution-results}
  \scriptsize
  \renewcommand{\arraystretch}{0.96}
  \setlength{\tabcolsep}{3pt}
  \newcommand{\change}[1]{%
    \unskip%
    \ifdim#1pt>0pt
      \textcolor{green!50!black}{\,(\ensuremath{#1}\%)}%
    \else
      \ifdim#1pt<0pt
        \textcolor{red!70!black}{\,(\ensuremath{#1}\%)}%
      \else
        \textcolor{yellow!50!black}{\,(\ensuremath{#1}\%)}%
      \fi
    \fi}
  \newcommand{\best}[1]{{\fontseries{b}\selectfont\color{black!92}#1}}
  \newcommand{\second}[1]{\underline{#1}}
  \resizebox{\linewidth}{!}{%
  \begin{tabular}{@{}lllcccc@{}}
    \toprule
    Benchmark & Agent & Method & Train split & Gate split & Sealed split & Overall \\
    \midrule
    \multirow{10}{*}{Terminal Bench 2}
      & \multirow{5}{*}{MiniSWE} & \textit{Seed} & 32.6 & 12.4 & 11.2 & 56.2 \\
      &                           & AHE          & \second{37.1} \change{+4.5}  & \second{12.4} \change{+0.0}  & \second{15.7} \change{+4.5} & \second{65.2} \change{+9.0} \\
      &                           & Hyperagents  & 36.0 \change{+3.4}  & \best{16.9} \change{+4.5}  & \best{16.9} \change{+5.6} & \best{69.7} \change{+13.5} \\
      &                           & A-Evolve     & \best{40.5} \change{+7.9}  & \second{12.4} \change{+0.0}  & \best{16.9} \change{+5.6} & \best{69.7} \change{+13.5} \\
      &                           & GEPA         & 36.0 \change{+3.4}  & \second{12.4} \change{+0.0}  & \second{15.7} \change{+4.5} & 64.0 \change{+7.8} \\
    \cmidrule(lr){2-7}
      & \multirow{5}{*}{Codex}   & \textit{Seed} & 39.3 & 15.7 & 14.6 & 69.7 \\
      &                           & AHE          & \best{42.7} \change{+3.4}  & 12.4 \change{-3.4} & \best{16.9} \change{+2.2} & \best{71.9} \change{+2.2} \\
      &                           & Hyperagents  & \second{41.6} \change{+2.3}  & \second{13.5} \change{-2.2} & \second{15.7} \change{+1.1} & \second{70.8} \change{+1.1} \\
      &                           & A-Evolve     & 40.5 \change{+1.1}  & \best{15.7} \change{+0.0} & 14.6 \change{+0.0} & \second{70.8} \change{+1.1} \\
      &                           & GEPA         & 39.3 \change{+0.0}  & \best{15.7} \change{+0.0} & 14.6 \change{+0.0} & 69.7 \change{+0.0} \\
    \midrule
    \multirow{10}{*}{$\tau^3$ Banking}
      & \multirow{5}{*}{MiniSWE} & \textit{Seed} & 12.4 & 6.2 & 9.3 & 27.8 \\
      &                           & AHE          & \second{16.5} \change{+4.1} & 7.2 \change{+1.0} & 5.2 \change{-4.1} & 28.9 \change{+1.1} \\
      &                           & Hyperagents  & \best{17.5} \change{+5.1} & \second{8.2} \change{+2.0} & \second{7.2} \change{-2.1} & \best{33.0} \change{+5.2} \\
      &                           & A-Evolve     & 12.4 \change{+0.0} & 6.2 \change{+0.0} & \best{9.3} \change{+0.0} & 27.8 \change{+0.0} \\
      &                           & GEPA         & 11.3 \change{-1.1} & \best{11.3} \change{+5.1} & \second{7.2} \change{-2.1} & \second{29.9} \change{+2.1} \\
    \cmidrule(lr){2-7}
      & \multirow{5}{*}{Codex}   & \textit{Seed} & 16.5 & 5.2 & 3.1 & 24.7 \\
      &                           & AHE          & \best{17.5} \change{+1.0} & \second{8.2} \change{+3.0} & \best{8.2} \change{+5.1} & \best{34.0} \change{+9.3} \\
      &                           & Hyperagents  & 15.5 \change{-1.0} & 6.2 \change{+1.0} & \second{7.2} \change{+4.1} & 28.9 \change{+4.2} \\
      &                           & A-Evolve     & 15.5 \change{-1.0} & 7.2 \change{+2.0} & 6.2 \change{+3.1} & 28.9 \change{+4.2} \\
      &                           & GEPA         & \second{16.5} \change{+0.0} & \best{9.3} \change{+4.1} & \second{7.2} \change{+4.1} & \second{33.0} \change{+8.3} \\
    \bottomrule
  \end{tabular}%
  }
\end{table}

\textbf{Overall performance.} The results show substantial gains, with their size depending on the benchmark, seed agent, and evolution method. On Terminal Bench 2, every method improves the MiniSWE baseline by 7.8--13.5 percentage points; Hyperagents and A-Evolve reach the highest overall score of 69.7. Gains are smaller from the stronger Codex baseline: AHE reaches 71.9 ($+2.2$ points), Hyperagents and A-Evolve each add 1.1 points, and GEPA preserves the baseline. The pattern changes on $\tau^3$ Banking. Hyperagents gives the largest MiniSWE improvement, from 27.8 to 33.0, whereas AHE performs best with Codex, raising the overall score from 24.7 to 34.0. GEPA is close behind at 33.0. These shifts indicate that method rankings are conditional on both the task environment and the starting agent.

\textbf{Task split-level behavior.} The split scores expose tradeoffs hidden by the overall metric. With MiniSWE on Terminal Bench 2, all methods improve the train and sealed splits, while only Hyperagents improves the gate split. A-Evolve obtains the largest train gain ($+7.9$ points) and ties Hyperagents on sealed performance, yet its gate score is unchanged. With Codex, AHE and Hyperagents improve train and sealed performance while decreasing gate performance by 3.4 and 2.2 points, respectively. On $\tau^3$ Banking, Hyperagents produces the largest overall gain from MiniSWE despite a 2.1-point sealed regression; AHE and GEPA also lose 4.1 and 2.1 points on sealed tasks. The Codex results are more consistent across held-out splits: AHE improves train, gate, and sealed scores by 1.0, 3.0, and 5.1 points, while GEPA preserves train performance and improves both gate and sealed scores by 4.1 points.

\textbf{Implications.} No method dominates across benchmarks, seeds, and splits. These setting-dependent outcomes motivate \tool's support for changing evolution policies while retaining a common execution and evaluation pipeline. Gate regressions on Terminal Bench 2 and sealed regressions on $\tau^3$ Banking show why inspectable evidence matters: a higher overall score can conceal held-out losses. Split-level results tied to candidate records help researchers diagnose unsuccessful adaptations and choose which policies to investigate further. Substantial gains occur in several configurations, including on sealed tasks, but neither improvement nor transfer is uniform. These descriptive point estimates do not establish the stability of close differences without repeated runs.

\subsection{Execution fidelity}
\label{sec:capability-preservation}

\textbf{Protocol.} To test the fidelity conjecture without confounding it with evolution, we compare framework-mediated execution of an unchanged seed agent with Direct execution. Direct execution denotes invoking the frozen evaluator outside \tool without creating or recording framework state. Direct-A and Direct-B are two independent direct executions placed before and after the \tool execution, respectively. For each condition, we conduct three repetitions, each running the same unchanged seed agent in the fixed order Direct-A, \tool generation zero, and Direct-B under identical model, task, evaluator, and runtime settings. We report Avg@3 as the mean pass@1 across the three repetitions. Within each repetition, the Direct score is the mean of Direct-A and Direct-B. D--D agreement measures agreement between Direct-A and Direct-B. D--R agreement is the mean of the Direct-A--\tool and Direct-B--\tool task-level agreements, and excess disagreement is the mean cross-mode disagreement rate minus the D--D disagreement rate.

\begin{table}[H]
  \centering
  \caption{Capability preservation across three repetitions. Avg@3 is mean pass@1; D--D is Direct-A--Direct-B agreement, and D--R averages Direct-A--\tool and Direct-B--\tool agreement.}
  \label{tab:capability-preservation}
  \scriptsize
  \setlength{\tabcolsep}{3pt}
  \resizebox{0.8\linewidth}{!}{%
  \begin{tabular}{@{}llcccccc@{}}
    \toprule
    \multirow{2}{*}{Benchmark} & \multirow{2}{*}{Agent} & \multicolumn{3}{c}{Avg@3} & \multicolumn{2}{c}{Agreement} & \multirow{2}{*}{\shortstack{Excess\\disagreement}} \\
    \cmidrule(lr){3-5}\cmidrule(lr){6-7}
    & & Direct & \tool & $\Delta$ & D--D & D--R & \\
    \midrule
    Terminal Bench 2 & MiniSWE & 59.6 & 56.9 & $-2.6$ & 83.5 & 84.6 & $-1.1$ \\
    Terminal Bench 2 & Codex   & 64.0 & 61.4 & $-2.6$ & 88.0 & 82.0 & $+6.0$ \\
    $\tau^3$ Banking & MiniSWE & 24.6 & 22.0 & $-2.6$ & 81.1 & 80.9 & $+0.2$ \\
    $\tau^3$ Banking & Codex   & 29.6 & 33.0 & $+3.4$ & 82.1 & 80.1 & $+2.1$ \\
    \bottomrule
  \end{tabular}%
  }
\end{table}

\textbf{Results.} On Terminal Bench 2, both agents change by $-2.6$ points. MiniSWE's D--R agreement is 84.6\%, compared with 83.5\% D--D agreement. Codex has 82.0\% D--R agreement versus 88.0\% D--D agreement, yielding $+6.0$ points of excess disagreement; its RSIHub Avg@3 of 61.4 lies within the 60.7--65.2 range across the six direct executions. On $\tau^3$ Banking, MiniSWE and Codex change by $-2.6$ and $+3.4$ points, with excess disagreement of $+0.2$ and $+2.1$ points. Direct executions themselves disagree on 12.0--18.9\% of tasks. These controls characterize execution variability; they do not establish execution-mode equivalence or the statistical significance of the single-run evolution gains. See \Cref{sec:pairwise-aggre} for details.

\subsection{Recovery and integrity}
\label{sec:recovery-integrity}

\textbf{Protocol.} We test the integrity conjecture by terminating runs during mutation or evaluation and after writing an evaluation receipt, then resuming and comparing the recovered state with an uninterrupted control. We also attempt evaluator mutation, writes outside the declared mutable surface, access to gate, sealed, and anchor evidence, and forged archive outcomes. We run 30 independent trials per scenario in each benchmark campaign. A trial passes only if the workspace reaches a valid and attributable state, unauthorized changes cannot affect an accepted result, and the champion remains reproducible from framework evidence.

\begin{table}[H]
  \centering
  \caption{Fault-injection and boundary-enforcement results in two benchmark campaigns. Each correct-outcome entry is out of 30 trials; no trial admitted corrupted state. Latencies are reported separately rather than averaged across machines.}
  \label{tab:reliability-integrity}
  \scriptsize
  \renewcommand{\arraystretch}{1.08}
  \setlength{\tabcolsep}{3.2pt}
  \resizebox{\linewidth}{!}{%
  \begin{tabular}{@{}ll>{\raggedright\arraybackslash}m{5.4cm}cccc@{}}
    \toprule
    & & & \multicolumn{2}{c}{Terminal Bench 2} & \multicolumn{2}{c}{$\tau^3$ Banking} \\
    \cmidrule(lr){4-5}\cmidrule(lr){6-7}
    Class & Scenario & Required invariant & Correct & Median (s) & Correct & Median (s) \\
    \midrule
    Recovery & Terminate during mutation & Resume from a certified parent; admit no partial child & 30 & 4.72 & 30 & 5.15 \\
    Recovery & Terminate during evaluation & Preserve the old attempt and certify one terminal retry & 30 & 3.61 & 30 & 4.94 \\
    Recovery & Terminate after stamp/receipt & Reconcile receipts without a duplicate event & 30 & 3.27 & 30 & 4.81 \\
    Boundary & Modify evaluator files & Reject evaluator mutation before certified evaluation & 30 & 4.72 & 30 & 2.35 \\
    Boundary & Write outside the mutable surface & Exclude all out-of-surface edits from the candidate & 30 & 3.27 & 30 & 2.37 \\
    Boundary & Access protected evidence during mutation & Keep gate, sealed, and anchor canaries out of the mutation bundle & 30 & 0.23 & 30 & 1.63 \\
    Integrity & Forge an archive outcome & Fail closed on unstamped events or a forged champion cache & 30 & 0.98 & 30 & 0.69 \\
    \bottomrule
  \end{tabular}%
  }
\end{table}

\textbf{Results.} All 420 trials reached the required outcome without admitting corrupted state. Latencies are reported separately because the campaigns ran on different hosts. These results support integrity only within the tested lifecycle boundaries and attack cases. Method authors can reuse these tested recovery and evidence checks across recipes. \Cref{app:supplementary-evaluation} reports runtime overhead and method-integration effort.

\section{Conclusion}
\label{sec:conclusion}

\tool provides open, reusable infrastructure for agent evolution. All four evaluated adaptations reuse framework services without framework source changes. Gains and held-out transfer vary across settings, motivating shared protocols and inspectable evidence for policy comparison. We invite researchers to build on the released recipes, compare new policy components, and contribute operators that others can reuse.

The agent-driven path extends this separation to adaptive recipe revision: an outer agent may revise operator bindings and settings between iterations, while the evaluator, sealed split, and certified record remain protected. Together, these interfaces provide a shared foundation for studying agent self-improvement.

\section{Limitations}
\label{sec:limitations}

RSIHub's evidence guarantees rely on trusted framework services, host and
filesystem protections, Git and digest operations, configured external
services, and valid evaluators. Our results apply to the workloads, hosts, and
lifecycle boundaries tested in this study, and the capability-preservation
measurements assess task-level agreement rather than full behavioral
equivalence. Runtime fingerprints do
not replace host or service attestation, and RSIHub does not contain hostile
code, so stronger deployments require independent attestation, isolation, and
containment.

The reported results characterize fixed, recipe-driven runs rather
than adaptive recipe revision by an outer agent.

% Shared with the main paper.
\section{Responsible Use Statement}
% \section{Broader Impacts and Ethical Considerations}
\label{sec:responsible-use}

RSIHub can enable unsafe capability growth, reward gaming, or exposure of
sensitive traces. Higher pass@1 does not establish safe deployment or exclude
harmful intermediate actions; banking-benchmark success does not establish
suitability for financial decisions. RSIHub supports auditing and rollback,
not containment: editable operators and candidate adapters may execute on the
host, beyond task-container isolation and feedback filtering.

Deployers should restrict filesystem and network access, exclude production
credentials, and require held-out safety evaluation and human review before
promotion. Resource budgets and halt and rollback controls are needed, although
rollback cannot undo external actions or disclosures. Trajectories and archives
need access controls, retention limits, and redaction before sharing.

\clearpage

\bibliographystyle{plainnat}
\bibliography{references}

\begin{thebibliography}{44}
\providecommand{\natexlab}[1]{#1}
\providecommand{\url}[1]{\texttt{#1}}
\expandafter\ifx\csname urlstyle\endcsname\relax
  \providecommand{\doi}[1]{doi: #1}\else
  \providecommand{\doi}{doi: \begingroup \urlstyle{rm}\Url}\fi

\bibitem[Agrawal et~al.(2026)Agrawal, Tan, Soylu, Ziems, Khare, Opsahl-Ong,
  Singhvi, Shandilya, Ryan, Jiang, Potts, Sen, Dimakis, Stoica, Klein, Zaharia,
  and Khattab]{agrawal2025gepa}
Lakshya~A. Agrawal, Shangyin Tan, Dilara Soylu, Noah Ziems, Rishi Khare, Krista
  Opsahl-Ong, Arnav Singhvi, Herumb Shandilya, Michael~J. Ryan, Meng Jiang,
  Christopher Potts, Koushik Sen, Alex Dimakis, Ion Stoica, Dan Klein, Matei
  Zaharia, and Omar Khattab.
\newblock {GEPA}: Reflective prompt evolution can outperform reinforcement
  learning.
\newblock In \emph{International Conference on Learning Representations}, 2026.
\newblock URL
  \url{https://proceedings.iclr.cc/paper_files/paper/2026/hash/0e9e708b6f48e14fd0ac29e167413f76-Abstract-Conference.html}.

\bibitem[Cai et~al.(2026)Cai, Ziakas, Kontonis, Pearce, Sen, Krishnamurthy,
  Garg, and Papailiopoulos]{cai2026forkandflush}
Ziyang Cai, Christos Ziakas, Vasilis Kontonis, Tim Pearce, Siddhartha Sen,
  Akshay Krishnamurthy, Shivam Garg, and Dimitris Papailiopoulos.
\newblock {Fork-and-Flush}: Escaping idea basins in autoresearch agents.
\newblock \emph{arXiv preprint arXiv:2610.07447}, 2026.
\newblock \doi{10.48550/arXiv.2610.07447}.
\newblock URL \url{https://arxiv.org/abs/2610.07447}.

\bibitem[Chirigati et~al.(2016)Chirigati, Rampin, Shasha, and
  Freire]{chirigati2016reprozip}
Fernando Chirigati, R{\'e}mi Rampin, Dennis Shasha, and Juliana Freire.
\newblock {ReproZip}: Computational reproducibility with ease.
\newblock In \emph{Proceedings of the 2016 International Conference on
  Management of Data}, pages 2085--2088, 2016.
\newblock \doi{10.1145/2882903.2899401}.
\newblock URL \url{https://doi.org/10.1145/2882903.2899401}.

\bibitem[Denison et~al.(2024)Denison, MacDiarmid, Barez, Duvenaud, Kravec,
  Marks, Schiefer, Soklaski, Tamkin, Kaplan, Shlegeris, Bowman, Perez, and
  Hubinger]{denison2024subterfuge}
Carson Denison, Monte MacDiarmid, Fazl Barez, David Duvenaud, Shauna Kravec,
  Samuel Marks, Nicholas Schiefer, Ryan Soklaski, Alex Tamkin, Jared Kaplan,
  Buck Shlegeris, Samuel~R. Bowman, Ethan Perez, and Evan Hubinger.
\newblock Sycophancy to subterfuge: Investigating reward-tampering in large
  language models.
\newblock \emph{arXiv preprint arXiv:2406.10162}, 2024.
\newblock \doi{10.48550/arXiv.2406.10162}.
\newblock URL \url{https://arxiv.org/abs/2406.10162}.

\bibitem[Fernando et~al.(2024)Fernando, Banarse, Michalewski, Osindero, and
  Rockt{\"a}schel]{fernando2024promptbreeder}
Chrisantha Fernando, Dylan~Sunil Banarse, Henryk Michalewski, Simon Osindero,
  and Tim Rockt{\"a}schel.
\newblock Promptbreeder: Self-referential self-improvement via prompt
  evolution.
\newblock In \emph{Proceedings of the 41st International Conference on Machine
  Learning}, volume 235 of \emph{Proceedings of Machine Learning Research},
  pages 13481--13544, 2024.
\newblock URL \url{https://proceedings.mlr.press/v235/fernando24a.html}.

\bibitem[Gao et~al.(2023)Gao, Schulman, and Hilton]{gao2022overoptimization}
Leo Gao, John Schulman, and Jacob Hilton.
\newblock Scaling laws for reward model overoptimization.
\newblock In \emph{Proceedings of the 40th International Conference on Machine
  Learning}, volume 202 of \emph{Proceedings of Machine Learning Research},
  pages 10835--10866, 2023.
\newblock URL \url{https://proceedings.mlr.press/v202/gao23h.html}.

\bibitem[{Harbor Framework Team}(2026)]{harbor2026framework}
{Harbor Framework Team}.
\newblock {Harbor}: A framework for evaluating and optimizing agents and models
  in container environments.
\newblock Software, version 0.18.0, 2026.
\newblock URL \url{https://doi.org/10.5281/zenodo.21249820}.

\bibitem[Hu et~al.(2025)Hu, Lu, and Clune]{hu2024adas}
Shengran Hu, Cong Lu, and Jeff Clune.
\newblock Automated design of agentic systems.
\newblock In \emph{International Conference on Learning Representations}, 2025.
\newblock URL
  \url{https://proceedings.iclr.cc/paper_files/paper/2025/hash/36b7acf6f6010652b3f2a433774a66fe-Abstract-Conference.html}.

\bibitem[Jia et~al.(2025)Jia, Li, Kang, Wang, Wu, Wang, Wang, Zhang, Shen, Li,
  Qi, Liang, He, Zheng, and Zhu]{jia2025aihippocampus}
Zixia Jia, Jiaqi Li, Yipeng Kang, Yuxuan Wang, Tong Wu, Quansen Wang, Xiaobo
  Wang, Shuyi Zhang, Junzhe Shen, Qing Li, Siyuan Qi, Yitao Liang, Di~He,
  Zilong Zheng, and Song{-}Chun Zhu.
\newblock The {AI} hippocampus: How far are we from human memory?
\newblock \emph{Trans. Mach. Learn. Res.}, 2025, 2025.
\newblock URL \url{https://openreview.net/forum?id=Sk7pwmLuAY}.

\bibitem[Jimenez et~al.(2024)Jimenez, Yang, Wettig, Yao, Pei, Press, and
  Narasimhan]{jimenez2024swebench}
Carlos~E. Jimenez, John Yang, Alexander Wettig, Shunyu Yao, Kexin Pei, Ofir
  Press, and Karthik Narasimhan.
\newblock {SWE-bench}: Can language models resolve real-world {GitHub} issues?
\newblock In \emph{International Conference on Learning Representations}, 2024.
\newblock URL
  \url{https://proceedings.iclr.cc/paper_files/paper/2024/hash/edac78c3e300629acfe6cbe9ca88fb84-Abstract-Conference.html}.

\bibitem[Khattab et~al.(2024)Khattab, Singhvi, Maheshwari, Zhang, Santhanam, A,
  Haq, Sharma, Joshi, Moazam, Miller, Zaharia, and Potts]{khattab2023dspy}
Omar Khattab, Arnav Singhvi, Paridhi Maheshwari, Zhiyuan Zhang, Keshav
  Santhanam, Sri~Vardhamanan A, Saiful Haq, Ashutosh Sharma, Thomas Joshi,
  Hanna Moazam, Heather Miller, Matei Zaharia, and Christopher Potts.
\newblock {DSPy}: Compiling declarative language model calls into
  state-of-the-art pipelines.
\newblock In \emph{International Conference on Learning Representations}, 2024.
\newblock URL
  \url{https://proceedings.iclr.cc/paper_files/paper/2024/hash/f1cf02ce09757f57c3b93c0db83181e0-Abstract-Conference.html}.

\bibitem[Li et~al.(2024)Li, Wang, Ding, Wang, Kang, Jia, and Zheng]{li2024ram}
Jiaqi Li, Xiaobo Wang, Wentao Ding, Zihao Wang, Yipeng Kang, Zixia Jia, and
  Zilong Zheng.
\newblock {RAM:} towards an ever-improving memory system by learning from
  communications.
\newblock \emph{CoRR}, abs/2404.12045, 2024.
\newblock \doi{10.48550/ARXIV.2404.12045}.
\newblock URL \url{https://doi.org/10.48550/arXiv.2404.12045}.

\bibitem[Li et~al.(2025)Li, Dong, Liu, Yang, Wang, Wang, Zhu, Jia, and
  Zheng]{li2025reflectevo}
Jiaqi Li, Xinyi Dong, Yang Liu, Zhizhuo Yang, Quansen Wang, Xiaobo Wang,
  Song{-}Chun Zhu, Zixia Jia, and Zilong Zheng.
\newblock Reflectevo: Improving meta introspection of small llms by learning
  self-reflection.
\newblock In \emph{Findings of the Association for Computational Linguistics,
  {ACL} 2025, Vienna, Austria, July 27 - August 1, 2025}, volume {ACL} 2025 of
  \emph{Findings of {ACL}}, pages 16948--16966. Association for Computational
  Linguistics, 2025.
\newblock \doi{10.18653/V1/2025.FINDINGS-ACL.871}.
\newblock URL \url{https://doi.org/10.18653/v1/2025.findings-acl.871}.

\bibitem[Lin et~al.(2026{\natexlab{a}})Lin, Liu, Pan, Lin, Dou, Xi, Huang, Yan,
  Han, Gui, and Jiang]{lin2026ahe}
Jiahang Lin, Shichun Liu, Chengjun Pan, Lizhi Lin, Shihan Dou, Zhiheng Xi,
  Xuanjing Huang, Hang Yan, Zhenhua Han, Tao Gui, and Yu-Gang Jiang.
\newblock Agentic harness engineering: Observability-driven automatic evolution
  of coding-agent harnesses.
\newblock \emph{arXiv preprint arXiv:2604.25850}, 2026{\natexlab{a}}.
\newblock \doi{10.48550/arXiv.2604.25850}.
\newblock URL \url{https://arxiv.org/abs/2604.25850}.

\bibitem[Lin et~al.(2026{\natexlab{b}})Lin, Lu, Shi, He, Mao, Zhang, Wu, Tang,
  Liu, Dai, Zhang, Wang, Dumoulin, and Pei]{lin2026aevolve}
Minhua Lin, Hanqing Lu, Zhan Shi, Bing He, Rui Mao, Zhiwei Zhang, Zongyu Wu,
  Xianfeng Tang, Hui Liu, Zhenwei Dai, Xiang Zhang, Suhang Wang, Benoit
  Dumoulin, and Jian Pei.
\newblock Position: Agentic evolution is the path to evolving {LLMs}.
\newblock \emph{arXiv preprint arXiv:2602.00359}, 2026{\natexlab{b}}.
\newblock \doi{10.48550/arXiv.2602.00359}.
\newblock URL \url{https://arxiv.org/abs/2602.00359}.

\bibitem[Liu et~al.(2026)Liu, Su, Zhang, Guo, Xie, Jing, Xie, Zong, Yim, Zhang,
  Li, and Song]{liu2026rethinking}
Yuxuan Liu, Zhaochen Su, Yuhao Zhang, Jiahe Guo, Zhongwei Xie, Huihao Jing,
  Lingyun Xie, Qing Zong, Yauwai Yim, Zhixiong Zhang, Haoran Li, and Yangqiu
  Song.
\newblock Rethinking self-evolving agent skills: Feedback dynamics over
  multiple rounds.
\newblock \emph{arXiv preprint arXiv:2608.02636}, 2026.
\newblock \doi{10.48550/arXiv.2608.02636}.
\newblock URL \url{https://arxiv.org/abs/2608.02636}.

\bibitem[Madaan et~al.(2023)Madaan, Tandon, Gupta, Hallinan, Gao, Wiegreffe,
  Alon, Dziri, Prabhumoye, Yang, Gupta, Majumder, Hermann, Welleck,
  Yazdanbakhsh, and Clark]{madaan2023selfrefine}
Aman Madaan, Niket Tandon, Prakhar Gupta, Skyler Hallinan, Luyu Gao, Sarah
  Wiegreffe, Uri Alon, Nouha Dziri, Shrimai Prabhumoye, Yiming Yang, Shashank
  Gupta, Bodhisattwa~Prasad Majumder, Katherine Hermann, Sean Welleck, Amir
  Yazdanbakhsh, and Peter Clark.
\newblock {Self-Refine}: Iterative refinement with self-feedback.
\newblock In \emph{Advances in Neural Information Processing Systems},
  volume~36, pages 46534--46594, 2023.
\newblock \doi{10.52202/075280-2019}.
\newblock URL
  \url{https://papers.nips.cc/paper/2023/hash/91edff07232fb1b55a505a9e9f6c0ff3-Abstract-Conference.html}.

\bibitem[Merrill et~al.(2026)Merrill, Shaw, Carlini, Li, Raj, Bercovich, Shi,
  Shin, Walshe, Buchanan, Shen, Ye, Lin, Poulos, Wang, Nezhurina, Lu,
  Mastromichalakis, Xu, Chen, Liu, Zhang, Chen, Kashyap, Uslu, Li, Wu, Yan,
  Bian, Sharma, Sun, Dillmann, Anand, Lanpouthakoun, Koopah, Hu, Guha, Dreiman,
  Zhu, Krauth, Zhong, Muennighoff, Amanfu, Tan, Pimpalgaonkar, Aggarwal, Lin,
  Lan, Zhao, Liang, Wang, Wang, Zhou, Heineman, Liu, Trivedi, Yang, Lin,
  Shetty, Yang, Omi, Raoof, Li, Zhuo, Lin, Dai, Wang, Chai, Zhou, Wahdany, She,
  Hu, Dong, Zhu, Cui, Saiyed, Kolbeinsson, Rytting, Marten, Wang, Jitsev,
  Dimakis, Konwinski, and Schmidt]{merrill2026terminalbench}
Mike~A Merrill, Alexander~Glenn Shaw, Nicholas Carlini, Boxuan Li, Harsh Raj,
  Ivan Bercovich, Lin Shi, Jeong~Yeon Shin, Thomas Walshe, E.~Kelly Buchanan,
  Junhong Shen, Guanghao Ye, Haowei Lin, Jason Poulos, Maoyu Wang, Marianna
  Nezhurina, Di~Lu, Orfeas~Menis Mastromichalakis, Zhiwei Xu, Zizhao Chen, Yue
  Liu, Robert Zhang, Leon~Liangyu Chen, Anurag Kashyap, Jan-Lucas Uslu, Jeffrey
  Li, Jianbo Wu, Minghao Yan, Song Bian, Vedang Sharma, Ke~Sun, Steven
  Dillmann, Akshay Anand, Andrew Lanpouthakoun, Bardia Koopah, Changran Hu,
  Etash~Kumar Guha, Gabriel H.~S. Dreiman, Jiacheng Zhu, Karl Krauth, Li~Zhong,
  Niklas Muennighoff, Robert~Kwesi Amanfu, Shangyin Tan, Shreyas Pimpalgaonkar,
  Tushar Aggarwal, Xiangning Lin, Xin Lan, Xuandong Zhao, Yiqing Liang, Yuanli
  Wang, Zilong Wang, Changzhi Zhou, David Heineman, Hange Liu, Harsh Trivedi,
  John Yang, Junhong Lin, Manish Shetty, Michael Yang, Nabil Omi, Negin Raoof,
  Shanda Li, Terry~Yue Zhuo, Wuwei Lin, Yiwei Dai, Yuxin Wang, Wenhao Chai,
  Shang Zhou, Dariush Wahdany, Ziyu She, Jiaming Hu, Zhikang Dong, Yuxuan Zhu,
  Sasha Cui, Ahson Saiyed, Arinbj{\"o}rn Kolbeinsson, Christopher~Michael
  Rytting, Ryan Marten, Yixin Wang, Jenia Jitsev, Alex Dimakis, Andy Konwinski,
  and Ludwig Schmidt.
\newblock Terminal-bench: Benchmarking agents on hard, realistic tasks in
  command line interfaces.
\newblock In \emph{The Fourteenth International Conference on Learning
  Representations}, 2026.
\newblock URL \url{https://openreview.net/forum?id=a7Qa4CcHak}.

\bibitem[Newman et~al.(2022)Newman, Meyers, and
  Torres-Arias]{newman2022sigstore}
Zachary Newman, John~Speed Meyers, and Santiago Torres-Arias.
\newblock Sigstore: Software signing for everybody.
\newblock In \emph{Proceedings of the 2022 ACM SIGSAC Conference on Computer
  and Communications Security}, pages 2353--2367, 2022.
\newblock \doi{10.1145/3548606.3560596}.

\bibitem[Novikov et~al.(2025)Novikov, V{\~u}, Eisenberger, Dupont, Huang,
  Wagner, Shirobokov, Kozlovskii, Ruiz, Mehrabian, Kumar, See, Chaudhuri,
  Holland, Davies, Nowozin, Kohli, and Balog]{novikov2025alphaevolve}
Alexander Novikov, Ng{\^a}n V{\~u}, Marvin Eisenberger, Emilien Dupont, Po-Sen
  Huang, Adam~Zsolt Wagner, Sergey Shirobokov, Borislav Kozlovskii, Francisco
  J.~R. Ruiz, Abbas Mehrabian, M.~Pawan Kumar, Abigail See, Swarat Chaudhuri,
  George Holland, Alex Davies, Sebastian Nowozin, Pushmeet Kohli, and Matej
  Balog.
\newblock {AlphaEvolve}: A coding agent for scientific and algorithmic
  discovery.
\newblock \emph{arXiv preprint arXiv:2506.13131}, 2025.
\newblock \doi{10.48550/arXiv.2506.13131}.
\newblock URL \url{https://arxiv.org/abs/2506.13131}.

\bibitem[{OpenAI}(2026)]{openai_codex}
{OpenAI}.
\newblock Openai codex cli.
\newblock Software, version 0.143.0, 2026.
\newblock URL \url{https://github.com/openai/codex/releases/tag/rust-v0.143.0}.
\newblock Released 2026-07-08; accessed 2026-08-24.

\bibitem[Paszke et~al.(2019)Paszke, Gross, Massa, Lerer, Bradbury, Chanan,
  Killeen, Lin, Gimelshein, Antiga, Desmaison, K{\"o}pf, Yang, DeVito, Raison,
  Tejani, Chilamkurthy, Steiner, Fang, Bai, and Chintala]{paszke2019pytorch}
Adam Paszke, Sam Gross, Francisco Massa, Adam Lerer, James Bradbury, Gregory
  Chanan, Trevor Killeen, Zeming Lin, Natalia Gimelshein, Luca Antiga, Alban
  Desmaison, Andreas K{\"o}pf, Edward Yang, Zachary DeVito, Martin Raison,
  Alykhan Tejani, Sasank Chilamkurthy, Benoit Steiner, Lu~Fang, Junjie Bai, and
  Soumith Chintala.
\newblock {PyTorch}: An imperative style, high-performance deep learning
  library.
\newblock In \emph{Advances in Neural Information Processing Systems},
  volume~32, 2019.
\newblock URL
  \url{https://papers.nips.cc/paper_files/paper/2019/hash/bdbca288fee7f92f2bfa9f7012727740-Abstract.html}.

\bibitem[Peng et~al.(2025)Peng, Shi, Wang, Zhang, Shen, and Gu]{peng2025sweqa}
Weihan Peng, Yuling Shi, Yuhang Wang, Xinyun Zhang, Beijun Shen, and Xiaodong
  Gu.
\newblock {SWE-QA}: Can language models answer repository-level code questions?
\newblock \emph{arXiv preprint arXiv:2509.14635}, 2025.
\newblock \doi{10.48550/arXiv.2509.14635}.
\newblock URL \url{https://arxiv.org/abs/2509.14635}.

\bibitem[Shang et~al.(2025)Shang, Li, Zhao, Ma, Liu, Xu, and
  Li]{shang2025agentsquare}
Yu~Shang, Yu~Li, Keyu Zhao, Likai Ma, Jiahe Liu, Fengli Xu, and Yong Li.
\newblock {AgentSquare}: Automatic {LLM} agent search in modular design space.
\newblock In \emph{International Conference on Learning Representations}, 2025.
\newblock URL
  \url{https://proceedings.iclr.cc/paper_files/paper/2025/hash/0ae94013da7cd459402fd77874e09ee3-Abstract-Conference.html}.

\bibitem[Sheng et~al.(2025)Sheng, Zhang, Ye, Wu, Zhang, Zhang, Peng, Lin, and
  Wu]{sheng2025hybridflow}
Guangming Sheng, Chi Zhang, Zilingfeng Ye, Xibin Wu, Wang Zhang, Ru~Zhang,
  Yanghua Peng, Haibin Lin, and Chuan Wu.
\newblock {HybridFlow}: A flexible and efficient {RLHF} framework.
\newblock In \emph{Proceedings of the Twentieth European Conference on Computer
  Systems}, pages 1279--1297, 2025.
\newblock \doi{10.1145/3689031.3696075}.
\newblock URL \url{https://doi.org/10.1145/3689031.3696075}.

\bibitem[Shi et~al.(2026{\natexlab{a}})Shi, Zytek, Razavi, Narasimhan, and
  Barres]{shi2026tauknowledge}
Quan Shi, Alexandra Zytek, Pedram Razavi, Karthik Narasimhan, and Victor
  Barres.
\newblock {$\tau$-Knowledge}: Evaluating conversational agents over
  unstructured knowledge.
\newblock \emph{arXiv preprint arXiv:2603.04370}, 2026{\natexlab{a}}.
\newblock \doi{10.48550/arXiv.2603.04370}.
\newblock URL \url{https://arxiv.org/abs/2603.04370}.

\bibitem[Shi et~al.(2026{\natexlab{b}})Shi, Xu, Fu, Zeng, He, Zhang, Liu, Zhao,
  Zhuo, Cao, Ye, Liu, Cai, Cheung, and Gu]{shi2026swebenchpromax}
Yuling Shi, Jinghan Xu, Kelin Fu, Wenhao Zeng, Shilin He, Lei Zhang, Yue Liu,
  Zelin Zhao, Terry~Yue Zhuo, Jialun Cao, Siyu Ye, Tianyu Liu, Kai Cai,
  Shing-Chi Cheung, and Xiaodong Gu.
\newblock {SWE-Bench ProMax}: Benchmarking agents on large-scale multilingual
  code refactoring.
\newblock \emph{arXiv preprint arXiv:2608.09802}, 2026{\natexlab{b}}.
\newblock \doi{10.48550/arXiv.2608.09802}.
\newblock URL \url{https://arxiv.org/abs/2608.09802}.

\bibitem[Shinn et~al.(2023)Shinn, Cassano, Gopinath, Narasimhan, and
  Yao]{shinn2023reflexion}
Noah Shinn, Federico Cassano, Ashwin Gopinath, Karthik Narasimhan, and Shunyu
  Yao.
\newblock Reflexion: Language agents with verbal reinforcement learning.
\newblock In \emph{Advances in Neural Information Processing Systems},
  volume~36, pages 8634--8652, 2023.
\newblock \doi{10.52202/075280-0377}.
\newblock URL
  \url{https://papers.nips.cc/paper_files/paper/2023/hash/1b44b878bb782e6954cd888628510e90-Abstract-Conference.html}.

\bibitem[Torres-Arias et~al.(2019)Torres-Arias, Afzali, Kuppusamy, Curtmola,
  and Cappos]{torresarias2019intoto}
Santiago Torres-Arias, Hammad Afzali, Trishank~Karthik Kuppusamy, Reza
  Curtmola, and Justin Cappos.
\newblock {in-toto}: Providing farm-to-table guarantees for bits and bytes.
\newblock In \emph{28th USENIX Security Symposium}, pages 1393--1410, 2019.
\newblock URL
  \url{https://www.usenix.org/conference/usenixsecurity19/presentation/torres-arias}.

\bibitem[Wang et~al.(2025{\natexlab{a}})Wang, Jia, Li, Liu, and
  Zheng]{wang2025uapo}
Xiaobo Wang, Zixia Jia, Jiaqi Li, Qi~Liu, and Zilong Zheng.
\newblock Adaptive preference optimization with uncertainty-aware utility
  anchor.
\newblock In \emph{Findings of the Association for Computational Linguistics:
  {EMNLP} 2025, Suzhou, China, November 4-9, 2025}, pages 19204--19225.
  Association for Computational Linguistics, 2025{\natexlab{a}}.
\newblock \doi{10.18653/V1/2025.FINDINGS-EMNLP.1046}.
\newblock URL \url{https://doi.org/10.18653/v1/2025.findings-emnlp.1046}.

\bibitem[Wang et~al.(2026{\natexlab{a}})Wang, Wu, Tang, Li, Liu, and
  Zheng]{wang2026flip}
Xiaobo Wang, Tong Wu, Min Tang, Jiaqi Li, Qi~Liu, and Zilong Zheng.
\newblock The flip side of {RLHF:} on-policy feedback for reward model
  self-supervised improvement.
\newblock \emph{CoRR}, abs/2605.30888, 2026{\natexlab{a}}.
\newblock \doi{10.48550/ARXIV.2605.30888}.
\newblock URL \url{https://doi.org/10.48550/arXiv.2605.30888}.

\bibitem[Wang et~al.(2026{\natexlab{b}})Wang, Rosenberg, Michelini, Smith,
  Tran, Nyst, Malhotra, Zhou, Chen, Brennan, and Neubig]{openhandsplatform}
Xingyao Wang, Simon Rosenberg, Juan Michelini, Calvin Smith, Hoang~H. Tran,
  Engel Nyst, Rohit Malhotra, Xuhui Zhou, Valerie Chen, Robert Brennan, and
  Graham Neubig.
\newblock The {OpenHands} software agent {SDK}: A composable and extensible
  foundation for production agents.
\newblock In \emph{Proceedings of Machine Learning and Systems}, volume~8,
  2026{\natexlab{b}}.
\newblock URL
  \url{https://proceedings.mlsys.org/paper_files/paper/2026/hash/8ae9cf363ea625161f885b798c1f1f78-Abstract-Conference.html}.

\bibitem[Wang et~al.(2025{\natexlab{b}})Wang, Liu, Fang, and
  Meng]{wang2025evoagentx}
Yingxu Wang, Siwei Liu, Jinyuan Fang, and Zaiqiao Meng.
\newblock {EvoAgentX}: An automated framework for evolving agentic workflows.
\newblock In \emph{Proceedings of the 2025 Conference on Empirical Methods in
  Natural Language Processing: System Demonstrations}, pages 643--655, Suzhou,
  China, 2025{\natexlab{b}}. Association for Computational Linguistics.
\newblock \doi{10.18653/v1/2025.emnlp-demos.47}.
\newblock URL \url{https://aclanthology.org/2025.emnlp-demos.47/}.

\bibitem[Wang et~al.(2026{\natexlab{c}})Wang, Shi, Li, Liu, Zhang, Wan, and
  Gu]{wang2026effiskill}
Zimu Wang, Yuling Shi, Mengfan Li, Zijun Liu, Jie~M. Zhang, Chengcheng Wan, and
  Xiaodong Gu.
\newblock {EffiSkill}: Agent skill based automated code efficiency
  optimization.
\newblock \emph{arXiv preprint arXiv:2603.27850}, 2026{\natexlab{c}}.
\newblock \doi{10.48550/arXiv.2603.27850}.
\newblock URL \url{https://arxiv.org/abs/2603.27850}.

\bibitem[Wu et~al.(2026)Wu, Lyu, Luo, Wang, Ye, Lin, Ji, Jiang, Wang, Wang, Ye,
  Wang, Wang, Liu, Wang, Cai, Xiong, Fang, Liu, Zhang, Yang, Hu, and
  Liu]{wu2026ai4ai}
Kai Wu, Hao Lyu, Zhen Luo, Chaofan Wang, Siyu Ye, Jinghao Lin, Xiaozhong Ji,
  Boyuan Jiang, Shengzhi Wang, Zihan Wang, Yiwen Ye, Hao Wang, Zimu Wang,
  Wenzhe Liu, Ruobing Wang, Kai Cai, Mingliang Xiong, Wen Fang, Mingqing Liu,
  Yifan Zhang, Lei Yang, Xiaobin Hu, and Qingwen Liu.
\newblock {AI4AI} survey: From long-horizon agents to recursive
  self-improvement---definitions, reliable horizons, and open problems.
\newblock Preprints.org, 2026.
\newblock URL \url{https://www.preprints.org/manuscript/202608.2108/v1}.
\newblock Version 1. Preprint.

\bibitem[Yang et~al.(2024)Yang, Jimenez, Wettig, Lieret, Yao, Narasimhan, and
  Press]{yang2024sweagent}
John Yang, Carlos Jimenez, Alexander Wettig, Kilian Lieret, Shunyu Yao, Karthik
  Narasimhan, and Ofir Press.
\newblock {SWE}-agent: Agent-computer interfaces enable automated software
  engineering.
\newblock In \emph{Advances in Neural Information Processing Systems},
  volume~37, pages 50528--50652, 2024.
\newblock \doi{10.52202/079017-1601}.
\newblock URL
  \url{https://proceedings.neurips.cc/paper_files/paper/2024/hash/5a7c947568c1b1328ccc5230172e1e7c-Abstract-Conference.html}.

\bibitem[Yuksekgonul et~al.(2025)Yuksekgonul, Bianchi, Boen, Liu, Lu, Huang,
  Guestrin, and Zou]{yuksekgonul2024textgrad}
Mert Yuksekgonul, Federico Bianchi, Joseph Boen, Sheng Liu, Pan Lu, Zhi Huang,
  Carlos Guestrin, and James Zou.
\newblock Optimizing generative {AI} by backpropagating language model
  feedback.
\newblock \emph{Nature}, 639\penalty0 (8055):\penalty0 609--616, 2025.
\newblock \doi{10.1038/s41586-025-08661-4}.
\newblock URL \url{https://doi.org/10.1038/s41586-025-08661-4}.

\bibitem[Zelikman et~al.(2024)Zelikman, Lorch, Mackey, and
  Kalai]{zelikman2024stop}
Eric Zelikman, Eliana Lorch, Lester Mackey, and Adam~Tauman Kalai.
\newblock Self-taught optimizer {(STOP)}: Recursively self-improving code
  generation.
\newblock In \emph{Conference on Language Modeling}, 2024.
\newblock URL \url{https://openreview.net/forum?id=46Zgqo4QIU}.

\bibitem[Zhang et~al.(2026{\natexlab{a}})Zhang, Fan, Zou, Chen, Wang, Zhou, Li,
  Huang, Yao, Zheng, Liu, Li, and Yu]{zhang2026evoskills}
Hanrong Zhang, Shicheng Fan, Henry~Peng Zou, Yankai Chen, Zhenting Wang, Jiayu
  Zhou, Chengze Li, Wei-Chieh Huang, Yifei Yao, Kening Zheng, Xue~(Steve) Liu,
  Xiaoxiao Li, and Philip~S. Yu.
\newblock {CoEvoSkills}: Self-evolving agent skills via co-evolutionary
  verification.
\newblock \emph{arXiv preprint arXiv:2604.01687}, 2026{\natexlab{a}}.
\newblock \doi{10.48550/arXiv.2604.01687}.
\newblock URL \url{https://arxiv.org/abs/2604.01687}.
\newblock Accepted at the Conference on Language Modeling (COLM 2026).

\bibitem[Zhang et~al.(2026{\natexlab{b}})Zhang, Hu, Lu, Lange, and
  Clune]{zhang2025dgm}
Jenny Zhang, Shengran Hu, Cong Lu, Robert Lange, and Jeff Clune.
\newblock Darwin {G}{\"o}del machine: Open-ended evolution of self-improving
  agents.
\newblock In \emph{International Conference on Learning Representations},
  2026{\natexlab{b}}.
\newblock URL \url{https://iclr.cc/virtual/2026/poster/10007327}.

\bibitem[Zhang et~al.(2026{\natexlab{c}})Zhang, Zhao, Yang, Foerster, Clune,
  Jiang, Devlin, and Shavrina]{zhang2026hyperagents}
Jenny Zhang, Bingchen Zhao, Wannan Yang, Jakob Foerster, Jeff Clune, Minqi
  Jiang, Sam Devlin, and Tatiana Shavrina.
\newblock Hyperagents.
\newblock \emph{arXiv preprint arXiv:2603.19461}, 2026{\natexlab{c}}.
\newblock \doi{10.48550/arXiv.2603.19461}.
\newblock URL \url{https://arxiv.org/abs/2603.19461}.

\bibitem[Zhang et~al.(2026{\natexlab{d}})Zhang, Hu, Upasani, Ma, Hong,
  Kamanuru, Rainton, Wu, Ji, Li, Thakker, Zou, and Olukotun]{zhang2026ace}
Qizheng Zhang, Changran Hu, Shubhangi Upasani, Boyuan Ma, Fenglu Hong,
  Vamsidhar Kamanuru, Jay Rainton, Chen Wu, Mengmeng Ji, Hanchen Li, Urmish
  Thakker, James~Y. Zou, and Kunle Olukotun.
\newblock Agentic context engineering: Evolving contexts for self-improving
  language models.
\newblock In \emph{International Conference on Learning Representations},
  2026{\natexlab{d}}.
\newblock URL
  \url{https://proceedings.iclr.cc/paper_files/paper/2026/hash/8a94ff6f922d995d7d3f4ebf4143e442-Abstract-Conference.html}.

\bibitem[Zhang et~al.(2026{\natexlab{e}})Zhang, Wang, Liang, Shi, Zeng, Wang,
  He, Xu, Ye, Cai, and Gu]{zhang2026sweexplore}
Shaoqiu Zhang, Yuhang Wang, Jialiang Liang, Yuling Shi, Wenhao Zeng, Maoquan
  Wang, Shilin He, Ningyuan Xu, Siyu Ye, Kai Cai, and Xiaodong Gu.
\newblock {SWE-Explore}: Benchmarking how coding agents explore repositories.
\newblock \emph{arXiv preprint arXiv:2606.07297}, 2026{\natexlab{e}}.
\newblock \doi{10.48550/arXiv.2606.07297}.
\newblock URL \url{https://arxiv.org/abs/2606.07297}.

\bibitem[Zhuge et~al.(2024)Zhuge, Wang, Kirsch, Faccio, Khizbullin, and
  Schmidhuber]{zhuge2024gptswarm}
Mingchen Zhuge, Wenyi Wang, Louis Kirsch, Francesco Faccio, Dmitrii Khizbullin,
  and J{\"u}rgen Schmidhuber.
\newblock {GPTSwarm}: Language agents as optimizable graphs.
\newblock In \emph{Proceedings of the 41st International Conference on Machine
  Learning}, volume 235 of \emph{Proceedings of Machine Learning Research},
  pages 62743--62767, 2024.
\newblock URL \url{https://proceedings.mlr.press/v235/zhuge24a.html}.

\end{thebibliography}

\clearpage
\appendix
\section{Representative Recipe Configuration}
\label{app:recipe-configurations}

The listing below presents a representative GEPA-inspired recipe. Low-level
execution settings are omitted for brevity.

\begingroup
\scriptsize
\begin{verbatim}
experiment:
  max_generations: 10
  budget_usd: 150
  children_per_gen: 1
  seed: 0
target:
  seed: builtin-codex
surface:
  include: [target/**]
  exclude: []
operators:
  select:
    operator: pareto
    timeout_s: 600
    config:
      seed: 0
  rollout:
    operator: harbor
    timeout_s: 3600
    config:
      budget_tasks: 10
      task_sampling: generation_shuffle
  analyze:
    operator: gepa
    timeout_s: 600
    config:
      components:
        system_prompt: target/prompt.md
        task_execution_skill: target/skills/task-execution
      max_cases: 10
  mutate:
    operator: gepa
    timeout_s: 3600
    config:
      expose_gate_data: false
      runner: harbor
      agent: codex
      model: gpt-5.4
      agent_kwargs:
        reasoning_effort: xhigh
      editable_roots: [target]
      components:
        system_prompt: target/prompt.md
        task_execution_skill: target/skills/task-execution
      component_strategy: round_robin
      max_examples: 10
  validate:
    operator: minibatch_improvement
    timeout_s: 3600
    config:
      criterion: strict
      n_concurrent: 4
  gate:
    operator: parent_eligible
    timeout_s: 600
  record:
    operator: gepa
    timeout_s: 600
evaluator:
  engine: harbor
  dataset: terminal-bench@2.0
  agent: target.agent:HarborAgent
  split:
    train: 0.562
    gate: 0.213
    sealed: 0.225
    seed: 0
  sampling: static
  tasks_per_round: 10
  n_concurrent: 4
  repetitions: 1
  anchor:
    final: true
\end{verbatim}
\endgroup

\Needspace{24\baselineskip}
\section{Extended Evaluation}
\label{app:supplementary-evaluation}

\subsection{Pairwise Execution Agreement} \label{sec:pairwise-aggre}

\begin{table}[H]
  \centering
  \caption{Pairwise cross-mode task agreement over three repetitions. The mean D--R agreement is the arithmetic mean of the two pairwise agreements and is the value reported in \Cref{tab:capability-preservation}. All entries are percentages.}
  \label{tab:pairwise-fidelity}
  \small
  \setlength{\tabcolsep}{7pt}
  \begin{tabular}{@{}lcccc@{}}
    \toprule
    Benchmark & Agent & Direct-A--\tool & Direct-B--\tool & Mean D--R \\
    \midrule
    Terminal Bench 2 & MiniSWE & 84.3 & 85.0 & 84.6 \\
    Terminal Bench 2 & Codex   & 82.4 & 81.6 & 82.0 \\
    $\tau^3$ Banking & MiniSWE & 82.1 & 79.7 & 80.9 \\
    $\tau^3$ Banking & Codex   & 79.4 & 80.8 & 80.1 \\
    \bottomrule
  \end{tabular}
\end{table}

\textbf{Results.} On Terminal Bench 2, Direct-B--\tool agreement is 0.7 points higher for MiniSWE, whereas Direct-A--\tool agreement is 0.8 points higher for Codex. On $\tau^3$ Banking, Direct-A is 2.4 points higher for MiniSWE, while Direct-B is 1.4 points higher for Codex. Neither direct replicate is uniformly closer to framework-mediated execution, which motivates averaging the two comparisons.

\subsection{Runtime Overhead}

\textbf{Protocol.} We measured the cost of the services shared across recipes on both benchmark workloads by instrumenting workspace initialization, candidate snapshotting, mutable-surface validation, evaluation recording, integrity verification, and recovery separately from model inference, benchmark execution, and mutation. For Terminal Bench 2, initialization uses 50 microbenchmark repetitions, recovery uses 30 replay trials, and the four regular-operation rows and end-to-end shares use 20 complete Gen-1 runs, five per method. Each regular-operation sample sums the relevant exclusive spans within one generation, and we compute the median and P95 over the 20 per-generation totals. The reported generation share uses the union of the framework's exclusive spans rather than summing nested durations. Because the campaigns ran on different hosts, we report their latency distributions separately.

\begin{table}[H]
  \centering
  \caption{Framework overhead by benchmark campaign. Operation latencies are in milliseconds. On Terminal Bench 2, the framework's exclusive spans account for a median of 0.02\% and a P95 of 0.06\% of end-to-end generation time; on $\tau^3$ Banking, they account for 0.019\%.}
  \label{tab:framework-overhead}
  \small
  \setlength{\tabcolsep}{5.5pt}
  \begin{tabular}{@{}lcccc@{}}
    \toprule
    & \multicolumn{2}{c}{Terminal Bench 2} & \multicolumn{2}{c}{$\tau^3$ Banking} \\
    \cmidrule(lr){2-3}\cmidrule(lr){4-5}
    Framework operation & Median (ms) & P95 (ms) & Median (ms) & P95 (ms) \\
    \midrule
    Workspace initialization and configuration freeze & 155.5 & 368.8 & 1229.3 & 1320.4 \\
    Candidate worktree and snapshot creation             & 756.2 & 2952.3 & 19.2 & 21.6 \\
    Mutable-surface validation                            & 27.1 & 43.2 & 13.5 & 16.0 \\
    Evaluation stamp and archive update                  & 32.4 & 69.8 & 10.8 & 14.9 \\
    State verification and champion recomputation        & 652.5 & 1482.2 & 108.4 & 113.8 \\
    Recovery replay after interruption                   & 1006.8 & 2451.9 & 92.3 & 99.4 \\
    \bottomrule
  \end{tabular}
\end{table}

\textbf{Results.} Workspace initialization was the largest measured operation in the $\tau^3$ Banking campaign, with a median of 1.23 seconds, whereas candidate snapshot creation and state verification were the largest regular operations on Terminal Bench 2, with medians of 0.76 and 0.65 seconds. These absolute latencies should not be compared directly across the two hosts. More importantly, framework work remained small relative to complete generations: 0.019\% on $\tau^3$ Banking and a median of 0.02\% on Terminal Bench 2. Model, mutation, and benchmark execution therefore dominate runtime in both measured workloads.

\subsection{Method-Integration Effort}

\textbf{Protocol.} We evaluated whether the common interfaces reduce method-specific engineering by counting normalized recipe lines, new operator lines, reused operators, changes to framework source files, and integration tests for each adapted method. Reporting these quantities separately distinguishes declarative composition and reused services from genuinely new method logic.

\begin{table}[H]
  \centering
  \caption{Method-integration effort under the shared recipe and operator interfaces.}
  \label{tab:integration-effort}
  \scriptsize
  \setlength{\tabcolsep}{4pt}
  \begin{tabular}{@{}lccccc@{}}
    \toprule
    Method & Recipe LOC & New operator LOC & Reused operators & Framework files changed & Integration tests \\
    \midrule
    AHE         & 47 & 1087 & 2 & 0 & 47 \\
    Hyperagents & 48 & 216  & 2 & 0 & 15 \\
    A-Evolve    & 57 & 684  & 2 & 0 & 14 \\
    GEPA        & 61 & 745  & 2 & 0 & 16 \\
    \bottomrule
  \end{tabular}
\end{table}

\textbf{Results.} At the locked experiment commit, the four adaptations required 47--61 lines of recipe configuration and reused two operator bindings each. None required a change to a framework source file, while method-specific operator implementations ranged from 216 lines for Hyperagents to 1,087 lines for AHE. Each adaptation passed its 14--47 method-specific integration tests. The separation between zero framework changes and nonzero method-specific code supports the intended extension model: research logic remains in recipes and operators while candidate execution, evaluation, recovery, and evidence services are reused.

\Needspace{8\baselineskip}
\section{Inspecting an Execution}
\label{app:execution-viewer}

\begin{figure}[H]
  \centering
  \includegraphics[width=0.9\linewidth]{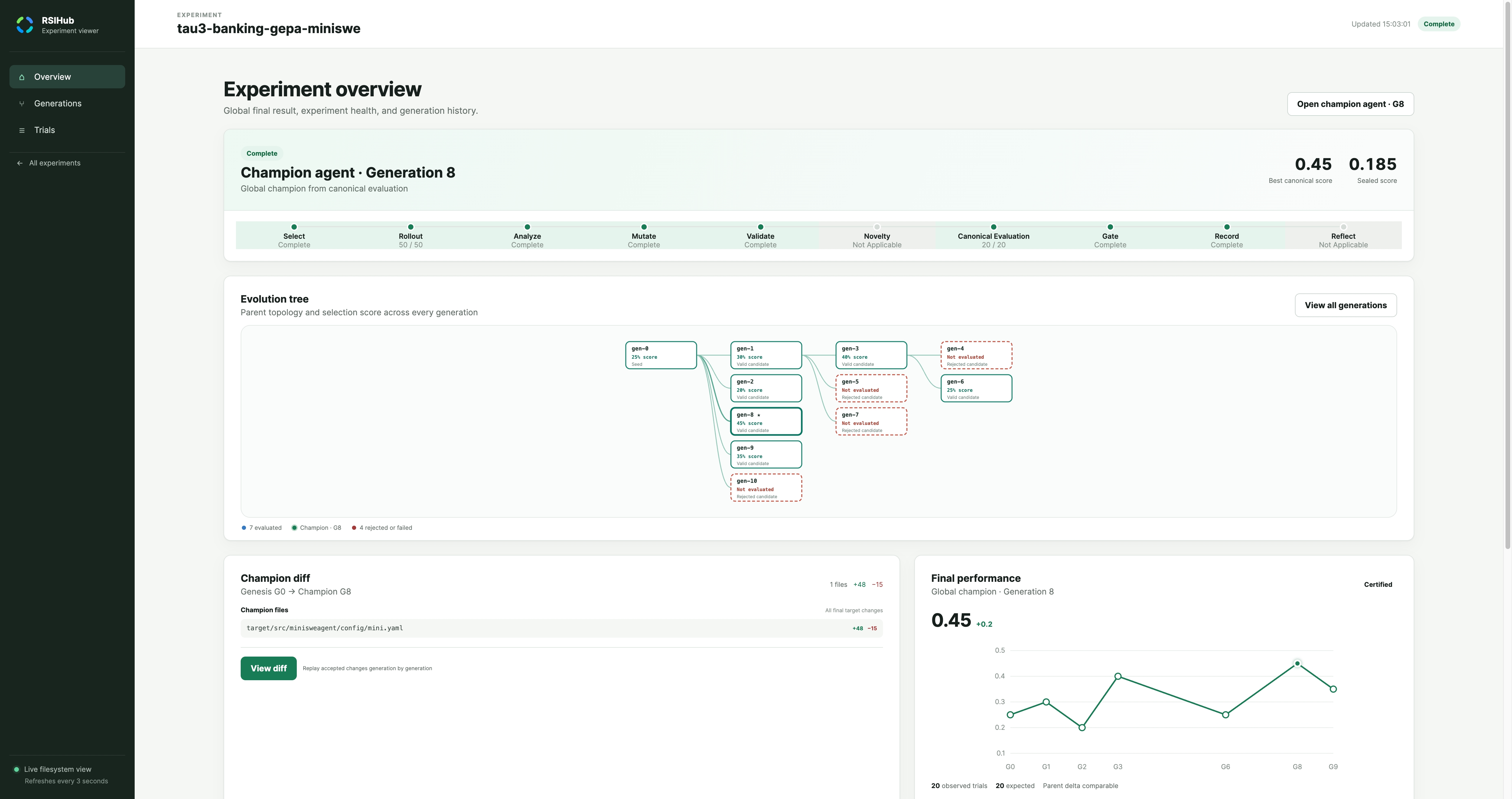}
  \caption{Inspecting one RSIHub execution. The read-only viewer identifies the
  accepted Generation~8 candidate as champion, shows its lifecycle status and
  run-specific primary-split and sealed scores, places it in the generation
  lineage, and summarizes its cumulative target-tree changes. The displayed
  values describe this execution; aggregate performance is reported in
  \Cref{sec:evaluation}.}
  \label{fig:viewer-walkthrough}
\end{figure}

RSIHub writes a file-backed experiment record that remains readable during and
after a run. Command-line summaries expose population state and integrity
findings. The framework identifies the champion as the best accepted candidate.
The read-only local viewer in \Cref{fig:viewer-walkthrough} presents experiment
status, generation lineage, and a summary of the champion's cumulative
target-tree changes.
In the illustrated execution, the accepted Generation~8 candidate is displayed
as the champion.

\Needspace{29\baselineskip}
\section{Executable Contracts and Recovery Semantics}
\label{app:contracts-recovery}

RSIHub exposes each method-controlled stage through a typed operator interface.
Operator outputs are validated before the framework
uses them; the framework separately enforces the mutable surface, commits the
candidate, runs protected evaluation, and protects framework record
fields. \Cref{tab:operator-contracts} summarizes the nine interfaces in
execution order.

\subsection{Lifecycle operators}

\begin{table}[H]
  \centering
  \caption{Lifecycle operator contracts. ``Req.'' denotes a registry-required
  operator and ``Opt.'' an operator enabled per recipe.}
  \label{tab:operator-contracts}
  \footnotesize
  \setlength{\tabcolsep}{4pt}
  \renewcommand{\arraystretch}{1.08}
  \renewcommand{\tabularxcolumn}[1]{>{\raggedright\arraybackslash}m{#1}}
  \begin{tabularx}{\linewidth}{@{}ll>{\raggedright\arraybackslash}m{0.30\linewidth}X@{}}
    \toprule
    Stage & Mode & Primary input & Validated result and framework handling \\
    \midrule
    Select & Req. & Archive view and execution context &
      Non-empty parent-generation list; the framework verifies eligibility and
      forks the recorded commit. \\
    Rollout & Req. & Candidate checkout and context &
      Structured summary and artifact paths retained as mutation evidence. \\
    Analyze & Opt. & Candidate checkout and context &
      Analysis summary and artifact paths bundled before mutation. \\
    Mutate & Req. & Checkout, observation text, and context &
      Changed paths, notes, and usage metadata; the framework checks the actual
      tree diff against the declared surface. \\
    Validate & Opt. & Mutated checkout and context &
      Acceptance flag, reason, and artifacts; rejection stops before protected evaluation. \\
    Novelty & Opt. & Uncommitted checkout and context &
      Finite novelty score and acceptance flag; rejection stops before commit. \\
    Gate & Req. & Evaluated child row, parent row, and context &
      \texttt{accept}/\texttt{reject} decision and reason; the framework keeps
      the score fields. \\
    Record & Req. & Terminal child row and context &
      Auxiliary annotation fields filtered against protected record fields. \\
    Reflect & Opt. & Updated archive and context &
      Validated append-only playbook operations; failure does not unwind the
      recorded generation. \\
    \bottomrule
  \end{tabularx}
\end{table}

\subsection{Evaluation identity and parent selection}

In strict-contract mode, an evaluation's contract accompanies its
scalar score. The contract binds the exact candidate to the evaluator, data, sampling,
runtime, and model conditions in \Cref{tab:evaluation-identity}; its
\texttt{contract\_id} is the canonical digest of the complete payload. Parent
selection combines the explicitly listed fixed-identity projection with
integrity and lifecycle checks; the full contract remains attached to each
attempt for provenance.

\begin{table}[H]
  \centering
  \caption{Evaluation identity and the parent-selection predicate.}
  \label{tab:evaluation-identity}
  \footnotesize
  \setlength{\tabcolsep}{4pt}
  \renewcommand{\arraystretch}{1.08}
  \renewcommand{\tabularxcolumn}[1]{>{\raggedright\arraybackslash}m{#1}}
  \begin{tabularx}{\linewidth}{@{}>{\raggedright\arraybackslash}m{0.20\linewidth}>{\raggedright\arraybackslash}m{0.39\linewidth}X@{}}
    \toprule
    Layer & Recorded identity & Role \\
    \midrule
    Candidate & Candidate commit and tree, generation, and optional dependency
      digest & Binds evidence to the evaluated source and relevant candidate
      dependencies. \\
    Evaluator & Frozen evaluator tree, semantic configuration digest, and
      framework version & Makes evaluator changes visible in per-attempt
      provenance. \\
    Data and trials & Dataset-content and task-set digests, ordered task
      members, split, repetitions, seed namespace, and trial identities & Makes
      task membership and multiplicity explicit. \\
    Runtime and model & Runtime digest, agent and model names, endpoint digest,
      concurrency, and retry policy & Records execution conditions that can
      affect evidence collection. \\
    Fixed comparison identity & Evaluator-tree fingerprint, task-set hash, and
      runtime fingerprint & Defines the cross-generation projection that a
      selectable parent must match. \\
    Integrity receipt & Framework marker and hash receipt for the serialized
      evaluation event & Flags an event when its framework's receipt is
      absent or no longer matches. \\
    Parent eligibility & Complete candidate or genesis evaluation;
      selection-eligible; no pending Gate decision; valid-parent and
      \texttt{keep}; certified receipt; matching fixed identity; finite score;
      and generation tag matching the recorded commit & Excludes incomplete,
      rejected, uncertified, incomparable, non-finite, and retagged candidates. \\
    \bottomrule
  \end{tabularx}
\end{table}

Sealed evaluation runs as a separate, non-selectable anchor. The implementation
stores it as auxiliary evidence, excludes it from the default mutation-feedback
bundle, and keeps its \texttt{anchor} purpose outside the candidate-or-genesis
parent predicate.

\subsection{Recovery checkpoints}

The archive is a recovery boundary: a complete certified candidate evaluation
is appended before Gate and remains pending until Gate records its decision. A
pending child is therefore unavailable to Select. Record annotations run after
Gate and are filtered separately; Record completion is outside the
parent-eligibility predicate. \Cref{tab:recovery-checkpoints} summarizes the
implemented restart behavior.

\begin{table}[H]
  \centering
  \caption{Checkpoint and recovery behavior.}
  \label{tab:recovery-checkpoints}
  \footnotesize
  \setlength{\tabcolsep}{4pt}
  \renewcommand{\arraystretch}{1.08}
  \renewcommand{\tabularxcolumn}[1]{>{\raggedright\arraybackslash}m{#1}}
  \begin{tabularx}{\linewidth}{@{}>{\raggedright\arraybackslash}m{0.24\linewidth}>{\raggedright\arraybackslash}m{0.39\linewidth}X@{}}
    \toprule
    Observed state & Recovery action & Preserved property \\
    \midrule
    Active child worktree at run start & Stop and require the operator to
      preserve edits or invoke repair. & The in-progress candidate remains
      untouched. \\
    Stale managed worktree & Repair removes only clean worktrees; dirty and
      externally managed worktrees are preserved and reported. & Cleanup is
      conservative with uncommitted work. \\
    Generation tag without a certified evaluation & Evaluate the existing tagged
      commit instead of repeating selection and mutation. & Resumption retains
      the created candidate's identity. \\
    Complete evaluation pending Gate & Resume Gate from the tagged candidate,
      then attempt terminal Record; keep the row ineligible until the Gate
      decision. & Parent selection waits for Gate finalization. \\
    Interrupted or terminally failed evaluation & Append available attempt
      evidence; require an explicit retry to create a new certified attempt. &
      Failed attempts remain auditable and retries are operator-controlled. \\
    Terminal Record already attempted & Skip a duplicate Record invocation on
      recovery. & Each terminal Record is invoked at most once. \\
    \bottomrule
  \end{tabularx}
\end{table}

% Code evidence: src/evolve/frozen/interfaces.py, src/evolve/frozen/sdk.py,
% src/evolve/driver.py, src/evolve/evaluation/contract.py,
% src/evolve/population.py, src/evolve/archive.py, and src/evolve/cli.py.

\end{document}